\documentclass[10pt,twocolumn,letterpaper]{article}

\usepackage[pagenumbers]{iccv} 

\usepackage{lineno}
\definecolor{iccvblue}{rgb}{0.21,0.49,0.74}
\usepackage[pagebackref,breaklinks,colorlinks,allcolors=iccvblue]{hyperref}

\def\paperID{6745} 
\def\confName{ICCV}
\def\confYear{2025}
\usepackage{multirow}
\usepackage{makecell}
\usepackage{colortbl}
\usepackage{tikz}
\usetikzlibrary{shapes.geometric}
\usetikzlibrary{fit}
\usetikzlibrary{decorations.pathreplacing}
\usetikzlibrary{positioning, shapes, arrows, arrows.meta, backgrounds}
\usepackage{bm}  
\usepackage{pgfplots}
\usepgfplotslibrary{colormaps} 
\pgfplotsset{compat=1.18}

\usepackage{fontawesome5} 
\usepackage{adjustbox}
\usepackage{arydshln}

\usepackage{pifont}
\newcommand{\cmark}{\ding{51}}%
\newcommand{\xmark}{\ding{55}}%
\usepackage{algorithm}
\usepackage{algpseudocode}
\usepackage{listings} 
\lstdefinestyle{PythonCommentStyle}{
    language=Python,
    basicstyle=\ttfamily\footnotesize,         
    commentstyle=\color{green!60!black},       
    keywordstyle=\color{blue!80!black},        
    stringstyle=\color{red!70!black},          
    showstringspaces=false,
    frame=none,
    xleftmargin=0pt,
    numbers=none,
    breaklines=true,                           
    breakindent=1em,                           
    linewidth=\columnwidth                     
}
\title{What Makes a Scene ? Scene Graph-based Evaluation and Feedback for Controllable Generation}
\author{Zuyao Chen\\
The Hong Kong Polytechnic University\\
{\tt\small zuyao.chen@connect.polyu.hk}
\and
Jinlin Wu, Zhen Lei\\
Institute of Automation, CAS\\
{\tt\small \{wujinlin2017, zhen.lei\}@ia.ac.cn}
\and 
Chang Wen Chen\\
The Hong Kong Polytechnic University\\
{\tt\small changwen.chen@polyu.edu.hk}
}
\begin{document}
\maketitle
\begin{abstract}
While text-to-image generation has been widely explored, 
synthesizing images from scene graphs remains relatively underexplored due to challenges in capturing complex spatial relationships and object interactions. 
To bridge this gap, we introduce Scene-Bench, a comprehensive benchmark for evaluating and enhancing factual consistency in natural scene generation.
Scene-Bench comprises MegaSG, a large-scale dataset of one million images annotated with detailed scene graphs, 
enabling extensive training and fair comparisons across diverse and intricate scenes. 
In addition, we propose SGScore, a novel evaluation metric that leverages the reasoning capabilities of multimodal large language models to assess both object presence and relationship accuracy, 
thereby providing a more precise measure of factual consistency than traditional metrics such as FID and CLIPScore. 
Furthermore, our scene graph feedback pipeline iteratively refines generated images by identifying and correcting discrepancies between the intended scene graph and the output. 
Extensive experiments demonstrate that Scene-Bench offers an effective evaluation framework for complex scene generation, 
and our feedback strategy significantly improves the factual consistency of image generation models, 
advancing the field of controllable image generation.
\end{abstract}
\section{Introduction}
\begin{figure}[t]
    \centering
    \includegraphics[width=\columnwidth]{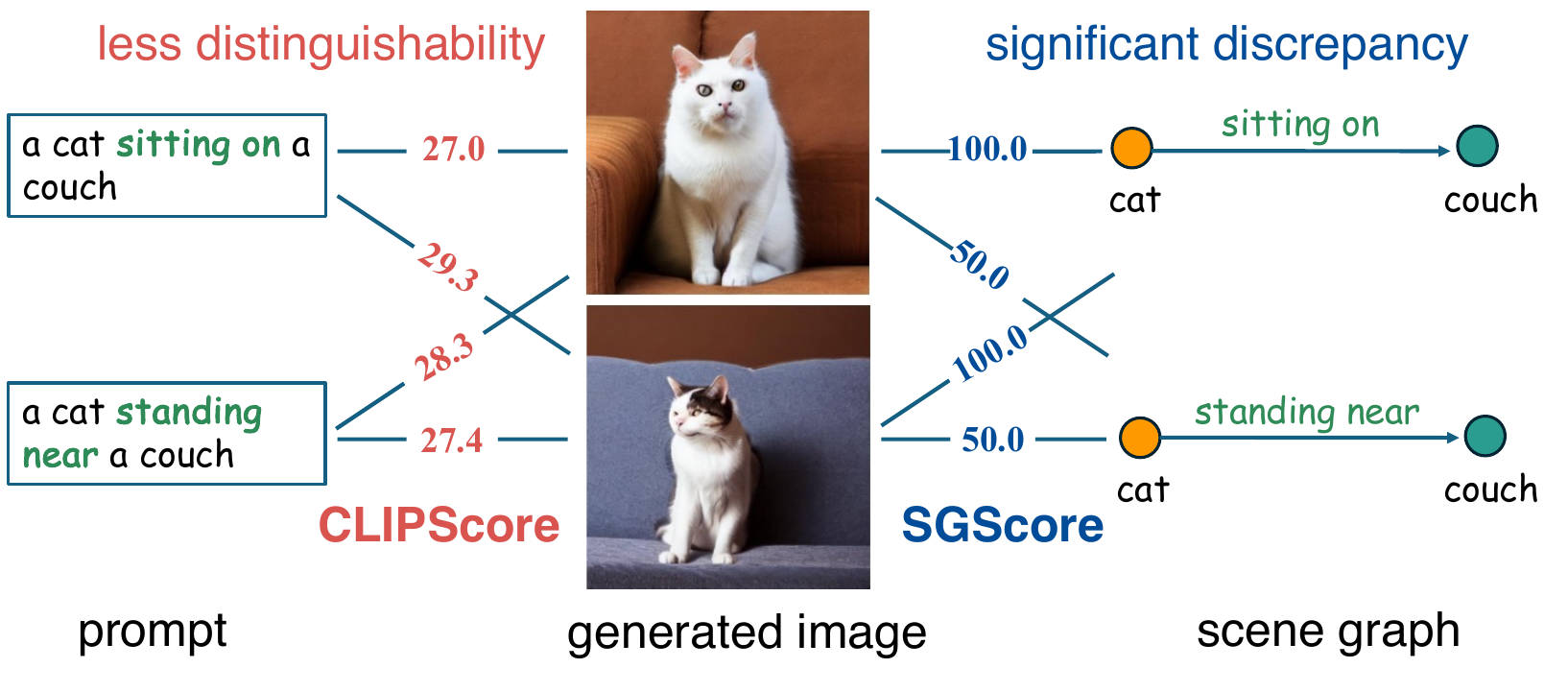}
    \caption{A comparison of CLIPScore~\cite{hessel2021clipscore} and the proposed SGScore for evaluating factual consistency. SGScore can distinguish such relationship discrepancies, while CLIPScore often overlooks them. 
    }
    \label{fig:comparison_clip_sgscore}
\end{figure}
\emph{``If you can't measure it, you can't improve it.''}
\begin{flushright}
-- Peter Drucker
\end{flushright}
 Generating realistic images coherent with natural scenes is important in numerous applications such as photo editing~\cite{isola2017image,zhu2017unpaired,kawar2023imagic}, content creation~\cite{karras2019style,brockds19large,rombach2022high}, \etc. 
Early generative models like Variational Autoencoders (VAE)~\cite{KingmaW13} produced blurry images due to limitations in modeling complex data distributions. 
Generative Adversarial Networks (GAN)~\cite{goodfellow2014generative} improved image quality but faced issues like training instability and mode collapse.
Recently, diffusion models like Stable Diffusion~\cite{rombach2022high} have proven effective for generating visually appealing images with realistic objects and high-resolution details~\cite{ho2020denoising,nichol2021glide,podellelbdmpr24,esser2024scaling}. 
Although diffusion models have achieved significant success, 
they still face challenges in generating complex scenes involving multiple objects \cite{yang2024mastering, liu2024draw}, 
particularly in ensuring factual consistency, 
such as the accurate presence of multiple objects and the correct relationship between objects within a natural image.

Recent efforts have aimed to address these limitations, focusing on compositional objects~\cite{liu2022compositional,feng2023training,lu2023tf}, improving text-image alignment~\cite{DBLP:journals/corr/abs-2302-12192,fan2024reinforcement}, and enhancing spatial consistency~\cite{ChatterjeeSAPGGSHLBY24}. 
For instance, methods like Composable Diffusion~\cite{liu2022compositional} compose multiple concepts by explicitly optimizing the defined energy functions, and Structured Diffusion~\cite{feng2023training} combines multiple objects by manipulating cross-attention layers. 
These methods improve the accuracy of multiple objects occurring in a single scene.
Additionally,  Chatterjee~\cite{ChatterjeeSAPGGSHLBY24} proposed a benchmark to evaluate and enhance the capability of modeling spatial relationships.

Despite this progress, \emph{how to evaluate the factual consistency between the condition (\eg, text, image, \etc) and the generated image} remains challenging. 
The difficulty lies in standard metrics like Fr\'{e}chet Inception Distance (FID)~\cite{heusel2017gans} and CLIPScore~\cite{hessel2021clipscore} primarily evaluate image quality but fall short in capturing factual consistency in complex scenes. 
FID, widely used to assess the visual fidelity of generated images, 
focuses on feature distribution matching between real and generated datasets. 
However, it overlooks spatial relationships and object interactions. 
For instance, images depicting a dog ``\emph{under a table}'' and ``\emph{on a table}'' may receive similar FID scores,
despite their vastly different relationships.
Similarly, CLIPScore measures semantic alignment between images and text by emphasizing global themes but cannot assess specific object relationships. 
CLIPScore may assign high scores to images that include all relevant objects but incorrectly depict their relationships, 
such as confusing ``\emph{a cat sitting on a couch}'' with ``\emph{a cat standing near a couch}'' (see \cref{fig:comparison_clip_sgscore}).
The drawback of FID and CLIPScore highlights the need for more specialized evaluation metrics to assess objects' presence and precise interactions between them.

To evaluate the factual consistency, we employ a structured representation known as a \emph{Scene Graph}, 
which has been demonstrated to outperform pure text in image retrieval~\cite{johnson2015image,krishna2017visual}. 
A scene graph encodes objects as nodes and their relationships as edges. 
For textual conditions, scene graphs can be parsed using natural language processing tools such as Scene Parser~\cite{mao2018parser} or large language models (LLMs). 
For image conditions, scene graphs are generated via Scene Graph Generation (SGG) models~\cite{xu2017scene, zellers2018neural, tang2019learning, chen2024expanding} or multimodal LLMs.
Leveraging this structured representation, we introduce a novel evaluation metric, \emph{SGScore}, 
which quantifies the factual consistency between generated images and their corresponding scene graphs. 
SGScore evaluates \emph{Object Recall} by verifying the presence of nodes and \emph{Relation Recall} by assessing the accuracy of edges within the scene graph. 
To adapt different domains and handle the extensive vocabulary inherent in generated images,
we utilize a multimodal LLM to perform these evaluations instead of relying on a pre-trained SGG model to convert images into scene graphs. 
Thanks to the reasoning and zero-shot capabilities of the multimodal LLM, 
SGScore can effectively distinguish between images depicting subtle differences, 
as shown in \cref{fig:comparison_clip_sgscore}.

When applying the new metric to benchmark different generative models, 
a critical bottleneck is the lack of large-scale datasets annotated with scene graphs, which are essential for fair and comprehensive comparisons. 
Existing datasets, such as Visual Genome (VG)~\cite{krishna2017visual} and COCO-Stuff~\cite{caesar2018coco}, are relatively small (\eg, only 5k and 2k images for testing, respectively), and their inherent long-tail distributions lead to biased evaluations. 
As a result, we develop \emph{MegaSG}, a large-scale dataset comprising one million images richly annotated with scene graphs that capture a wide range of objects and their complex relationships. 
MegaSG enables models to be trained and evaluated on diverse scenarios, from simple to highly intricate scenes, 
thus overcoming the limitations of previous datasets that were constrained by small-scale and biased distributions.

By combining the proposed \emph{SGScore} and \emph{MegaSG}, we introduce a novel benchmark, \emph{Scene-Bench}. 
To provide a comprehensive and fair benchmark, we sample images from MegaSG based on Scene Diversity and Scene Complexity. 
Scene Diversity sampling aims to evaluate model performance across diverse scene scenarios, while Scene Complexity sampling aims to evaluate model performance at different complexity levels. 
To our knowledge, \emph{Scene-Bench} is the first benchmark to evaluate generative models on a large-scale natural scene dataset using scene graphs.

Building upon this scene graph-based evaluation, we design a scene graph feedback pipeline that leverages multimodal LLMs for iterative refinement. 
The process begins with generating an initial image from a scene graph, followed by assessing factual consistency using the Scene-Bench metrics. 
When discrepancies are detected, such as missing objects or incorrect relationships, a missing graph is created to highlight these errors. 
A reference image is generated based on this missing graph to address the identified issues. 
By integrating this new image with the initial one, we refine the output, resulting in a final image that more accurately matches the intended scene described by the original scene graph.

In short, our contribution can be summarized as 
\begin{itemize}
    \item We introduce \emph{Scene-Bench}, a comprehensive and large-scale benchmark for evaluating factual consistency in scene graph-to-image generation. 
    \emph{Scene-Bench} includes \emph{MegaSG}, a dataset with one million images annotated with scene graphs, and a novel evaluation metric, \emph{SGScore}, which explicitly measures factual consistency by assessing the accuracy of object presence and relationships in generated images.
    \item We propose a scene graph feedback strategy that iteratively refines generated images by detecting and correcting discrepancies in object presence and relationship accuracy, thereby enhancing the factual consistency between the generated image and the intended scene.
    \item Extensive experiments demonstrate that \emph{Scene-Bench} provides a more comprehensive and effective evaluation benchmark for factual consistency in natural scenes. Furthermore, our proposed feedback pipeline significantly improves the factual consistency of generated images, particularly in complex scene scenarios.
\end{itemize}
\section{Related Work}
\textbf{Text-to-Image Generation (T2I).} 
The field of text-to-image generation has seen significant advancements with the transition from GANs~\cite{goodfellow2020generative} to diffusion models. 
Early GAN-based methods like StackGAN~\cite{zhang2017stackgan} and AttnGAN~\cite{xu2018attngan} generated images from textual descriptions but often struggled with image quality and diversity. 
The introduction of diffusion models marked a substantial improvement.
Models such as DALL-E~\cite{ramesh2021zero}, GLIDE~\cite{nichol2021glide}, and Stable Diffusion~\cite{rombach2022high} have achieved high-quality image synthesis with better text-image alignment by iteratively refining images from noise, conditioned on text prompts. 
Despite their success, these models face challenges in generating complex scenes involving multiple objects and ensuring consistency in object relationships.

\begin{figure*}[t] 
\definecolor{royalblue}{HTML}{004C99}
\definecolor{lightgray}{HTML}{E0E0E0}
\definecolor{personcolor}{HTML}{FF6666}  
\definecolor{ballcolor}{HTML}{66CCFF}    
    \centering
    \begin{minipage}{0.85\textwidth}
    \begin{subfigure}[t]{0.74\textwidth}
        \includegraphics[width=\textwidth]{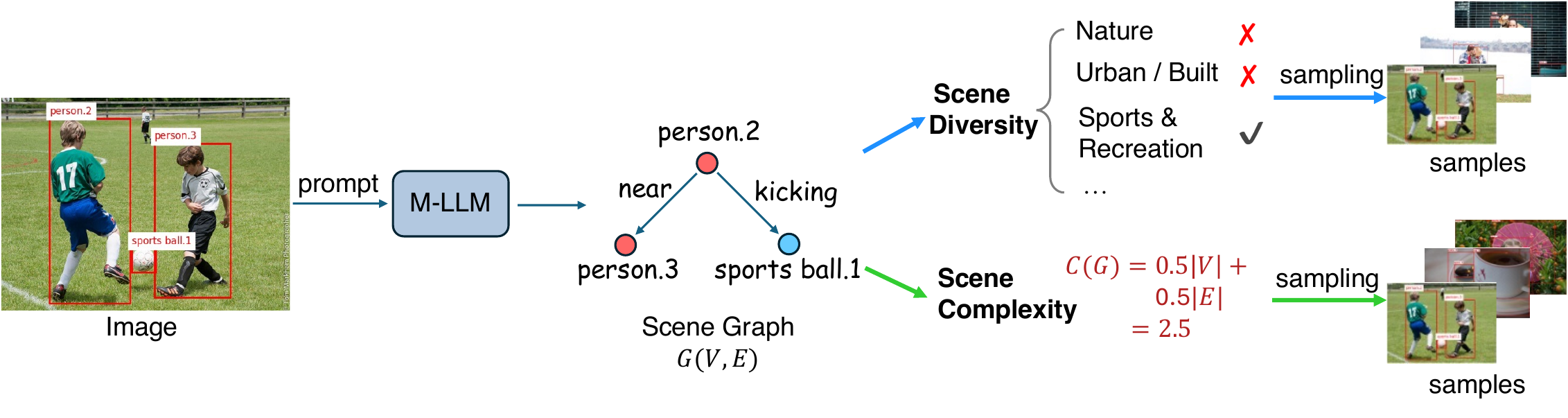}
        \subcaption{Creation of the MegaSG.}
    \end{subfigure}\hfill 
    \begin{subfigure}[t]{0.23\textwidth}
        \includegraphics[width=\textwidth]{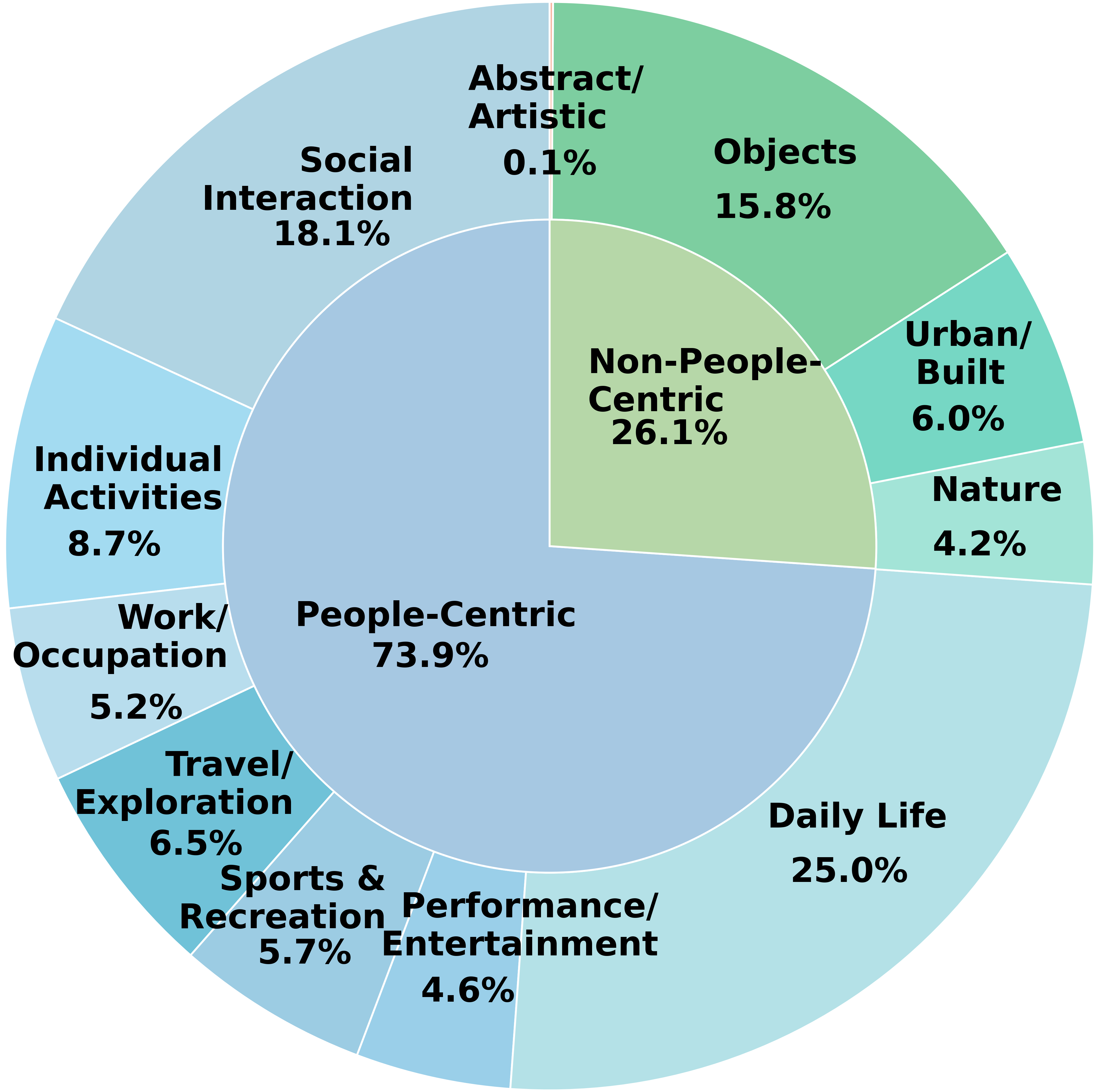}
        \subcaption{Scene distribution.}
    \end{subfigure}
    \end{minipage}
    \begin{subfigure}[t]{0.85\textwidth}
       \includegraphics[width=\textwidth]{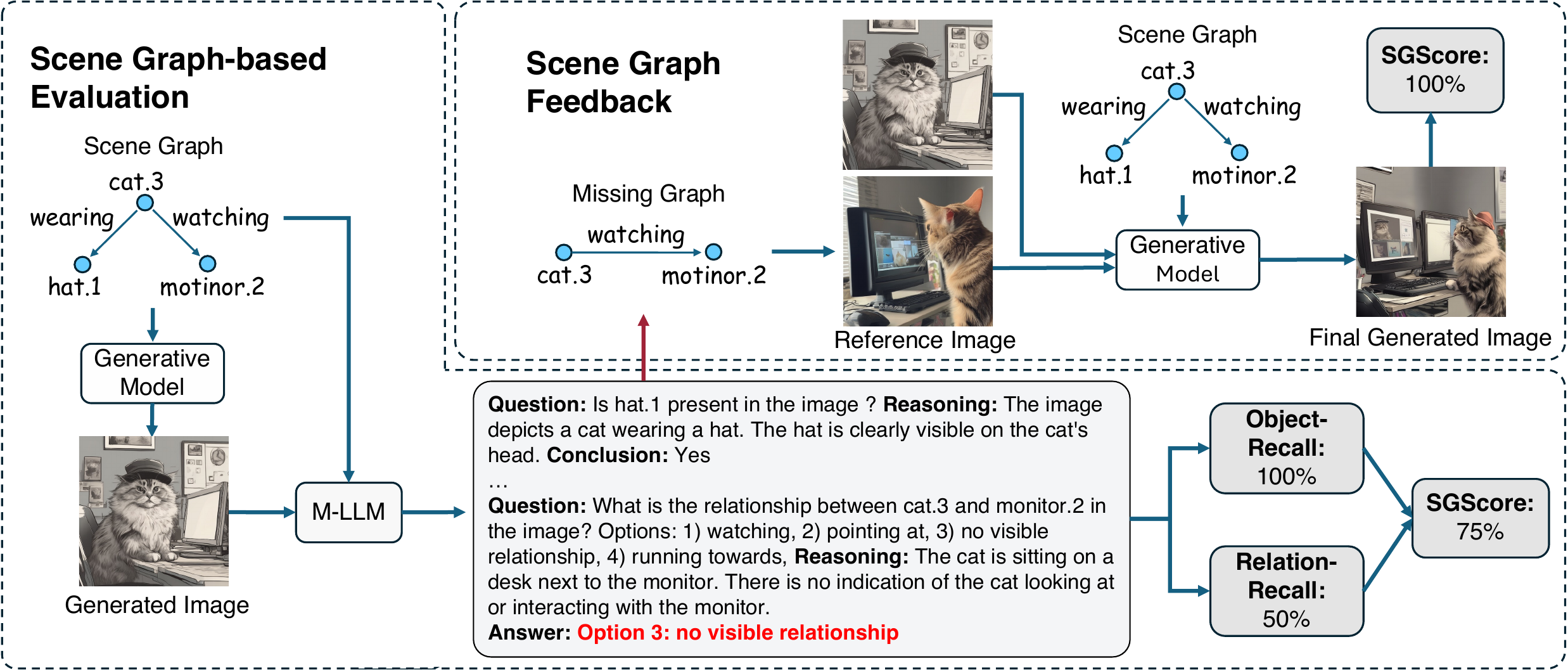}
       \subcaption{Scene Graph-based Evaluation and Scene Graph Feedback.}
    \end{subfigure}
    \caption{Overview of the Scene-Bench. (a) Scene graphs are generated from images using a multimodal LLM (M-LLM), capturing object relationships and interactions. Scene Diversity and Scene Complexity guide sampling to ensure dataset balance. (b) Scene distribution across categories highlights the diversity in People-Centric and Non-People-Centric themes. (c) Scene graph-based evaluation and feedback leverages the M-LLM to calculate object and relationship recall, generating an SGScore metric that quantifies factual consistency between the generated image and the intended scene. The feedback identifies and corrects discrepancies, iteratively refining the generated image.
    }
    \label{fig:scene-graph-example}
\end{figure*}
\noindent\textbf{Scene Graph-to-Image Generation (SG2IM).} Scene graphs offer a structured representation of objects and their relationships, providing a promising scheme for controllable image generation. 
Johnson \etal~\cite{johnson2018image} introduced SG2Im, a model that generates images from scene graphs using graph convolutional networks (GCN) and conditional GANs. 
Ashual and Wolf~\cite{ashual2019specifying} extended this approach by incorporating more detailed scene representations and object attributes. 
Recent methods have integrated scene graphs with diffusion models to enhance compositionality~\cite{liu2022compositional,farshad2023scenegenie,shen2024sg, liu2024r3cd, wang2024scene}. 
However, these approaches often require training on large-scale scene graph datasets and rely on additional guidance like bounding boxes(\eg, \cite{farshad2023scenegenie}) or specialized graph encoders (\eg, \cite{yang2022diffusion}),
which cannot be adapted to open-vocabulary scenarios. 
More importantly, evaluating these models remains challenging due to the lack of metrics that effectively capture scene-level fidelity,
including object presence and relationship accuracy in generated images.

\noindent\textbf{LLMs in Image Generation.} The integration of LLMs has opened new possibilities in image generation.
Recent works~\cite{feng2024layoutgpt,wu2024self,yang2024mastering,lian2023llm} have leveraged LLMs to enhance compositionality and controllability in image synthesis.
For example, LayoutGPT~\cite{feng2024layoutgpt} utilizes LLMs as visual planners to generate layouts from textual descriptions, improving user controllability. 
Similarly, methods like RPG~\cite{yang2024mastering} and Complex Diffusion~\cite{liu2024draw} leverage the reasoning capabilities of LLMs to decompose complex prompts into simpler tasks, aiding in the generation of complex scenes with multiple objects and relationships. 
However, their potential for providing feedback to iteratively refine scene graph-based generation has not been fully explored.

\noindent\textbf{Improving Relationship Consistency.} 
Addressing the limitations in capturing object relationships, 
several approaches have been proposed. 
Feng \etal~\cite{feng2023training} focused on improving compositional generalization in diffusion models through 
a modulated cross-attention mechanism.
Park \etal~\cite{park2021benchmark} introduced benchmarks specifically targeting compositional understanding in generative models. 
Chatterjee \etal~\cite{ChatterjeeSAPGGSHLBY24} designed a benchmark to evaluate the capability of modeling spatial relationships.
Despite these efforts, ensuring accurate depiction of relationships in complex scenes remains a significant challenge, and existing methods often do not provide mechanisms for iterative refinement based on explicit relationship feedback.
\begin{table}[t]
    \centering
    \resizebox{\columnwidth}{!}{
    \begin{tabular}{lccccc}
    \toprule
         \multirow{2}{*}{Dataset} &   \multirow{2}{*}{Images} &  \multirow{2}{*}{Obj./Rel.} & 
          \multicolumn{3}{c}{Test} 
         \\\cline{4-6}
         & & & Samples &   Triplets & Balanced \\ 
    \midrule 
         VG~\cite{krishna2017visual} &  108k & 179 / 49  & 5,096  & 12.8k & \textcolor[HTML]{D32F2F}{\xmark} \\ 
         COCO-Stuff~\cite{caesar2018coco} & 4.5k & 171 / 6  & 2,048
          & 22.7k &  \textcolor[HTML]{D32F2F}{\xmark}  \\
         MegaSG & 1M &  775 / 122  & 50,000 & 275k &  \textcolor[HTML]{1E90FF}{\cmark}
         \\  
        \bottomrule
    \end{tabular}
    }
\caption{Comparison of MegaSG with widely-used scene graph datasets. 
``Obj./Rel.'' denotes the number of object and relationship categories, 
while ``Balanced'' indicates whether the dataset is balanced with respect to scene complexity.
The VG statistics follow the pioneering work~\cite{johnson2018image} in SG2IM, and almost all subsequent works adopt this setting.
}
    \label{tab:dataset_comp}
\end{table}
\section{Scene-Bench}
Scene-Bench is a comprehensive benchmark that evaluates and enhances the factual consistency of natural scene generation from scene graphs by rigorously verifying both object presence and inter-object relationships.
Specifically, Scene-Bench consists of a large-scale dataset of scene graphs, 
and an autonomous evaluation pipeline. 
The overview of Scene-Bench is shown in \cref{fig:scene-graph-example}. 

\subsection{MegaSG: a large-scale dataset of scene graphs}
\label{sec:megasg}
\textbf{Creation of the Dataset.}
\label{sec:creation}
Due to the complexity and high cost of manual annotation, existing scene graph datasets, such as Visual Genome~\cite{krishna2017visual}, are relatively small in scale
(\eg, only 5k images are prepared for the test set \cite{johnson2018image}).
The limited size and inherent long-tail distribution make these datasets inadequate for studying diffusion models across diverse scene scenarios. 
To address this limitation and build a large-scale scene graph dataset, we leverage the reasoning capabilities of multimodal large language models (LLMs) in combination with pre-existing object detection datasets. 
Specifically, we collect {1 million} images from COCO~\cite{chen2015microsoft}, Object365~\cite{shao2019objects365}, and Open Images v6~\cite{kuznetsova2020open}, which offer rich object categories  and bounding boxes. 
These datasets are ideal for generating large-scale scene graphs efficiently.
For additional details, please see \cref{sec:supp_mega} of the \textbf{SMs} \footnote{Throught this paper, SMs refers to supplementary materials.}.


\noindent\textbf{Dataset Quality.}
To quantitatively verify the quality of MegaSG, we evaluate state-of-the-art Scene Graph Generation (SGG) models trained on different datasets. 
For instance, the OvSGTR (Swin-B)~\cite{chen2024expanding} model trained on MegaSG achieves a zero-shot performance recall of 45.71\% (R@50, PredCls mode) on the VG150 test set, outperforming models trained on smaller datasets like COCO Caption data (see \cref{tab:mega_zeroshot} of \textbf{SMs}). 
This improvement reflects the high-quality scene graph annotations of MegaSG, 
enabling further exploration of training SGG models on MegaSG or generating images from scene graphs on MegaSG.

\noindent\textbf{Scene Diversity.} 
To better understand the behavior of diffusion models in diverse scene scenarios, 
we utilized an LLM (\eg, Gemini 1.5 Flash~\cite{reid2024gemini}) to classify the MegaSG dataset into two main themes: \emph{People-Centric} and \emph{Non-People-Centric}.
The \emph{People-Centric} theme includes fine-grained categories such as \emph{Social Interaction}, \emph{Individual Activities}, \emph{Work / Occupation}, \emph{Travel / Exploration}, \emph{Sports \& Recreation}, \emph{Performance / Entertainment}, \emph{Daily Life}. 
For the \emph{Non-people-Centric} theme, we identified categories like \emph{Nature}, \emph{Urban / Built}, \emph{Objects}, and \emph{Abstract / Artistic}. 
The hierarchical distribution of these categories is illustrated in  \cref{fig:scene-graph-example} (b). 
And samples are shown in the figure (see \cref{fig:scene_cat} of \textbf{SMs}), 
showcasing the diversity and range of scenarios covered in the dataset.

\noindent\textbf{Scene Complexity.} 
In addition to categorizing natural scenes, 
measuring scene complexity is crucial for evaluating the performance of diffusion models. 
While simple scenes are generally easier for these models to handle, complex scenes pose greater challenges. 
This raises an important question: \emph{How can we quantitatively define the complexity of a natural scene?}

In this work, we define the complexity of a scene based on its scene graph $G = (V, E)$, where $V$ represents the set of nodes (objects) and $E$ represents the set of edges (relationships). 
The complexity is calculated as
\begin{equation}
    C(G) = \gamma \cdot |V| + (1.0 - \gamma) \cdot |E|, 
    \label{eq:comp}
\end{equation}
where $\gamma \in [0, 1]$ is a weighting factor that balances the influence of the number of nodes and edges.

The scene graph representation provides a straightforward way to quantify complexity, making it possible to analyze the performance of diffusion models across a range of difficulty levels, from simple to highly complex scenes. 
In contrast, defining the complexity of a text prompt is inherently more challenging due to the lack of explicit structural information.
By leveraging this graph-based approach, we can better understand how diffusion models respond to varying levels of scene complexity, offering insights into their strengths and limitations across different scenarios.

\noindent\textbf{Dataset Comparison.}
We compare the MegaSG with existing scene graph datasets, specifically Visual Genome (VG)~\cite{krishna2017visual} and COCO-Stuff~\cite{caesar2018coco}, as summarized in \cref{tab:dataset_comp}. 
MegaSG significantly outperforms VG and COCO-Stuff in terms of scale, encompassing 1 million images compared to VG's 108k and COCO-Stuff's 4.5k. 
Additionally, MegaSG offers a substantially richer vocabulary with 775 object categories and 122 relations, enhancing the diversity and complexity of scene annotations. 
Importantly, MegaSG's test set is \emph{complexity balanced}, ensuring an even distribution of simple, medium, and hard scenes, whereas VG and COCO-Stuff lack this balanced composition.

\subsection{Evaluation Strategy} 
\label{sec:evaluation}
To quantify the factual consistency,
we utilize a multimodal LLM (M-LLM) to assess the recall of objects and relationships,  
as shown in \cref{fig:scene-graph-example} (c). 

\noindent\textbf{Recall of Objects.} 
Given a generated image $I$ and its intended scene graph  $G = (V, E)$ , where $V$ represents the set of objects (nodes) and  $E$ represents the relationships (edges), we prompt the M-LLM with specific queries about the existence of each object.
For example, for a scene graph containing the relationships:
 \emph{\{``source": ``person.2", ``target": ``sports ball.1", ``relation": ``kicking"\}, \{``source": ``person.2", ``target": ``person.3", ``relation": ``near"\}},
we would prompt the M-LLM with questions such as ``\emph{Is there a sports ball in the image?}''.
The M-LLM, based on its multimodal capabilities, examines the generated image and responds with a binary answer (Yes / No) to indicate whether the specified object is present.
We define the object recall as the fraction of correctly identified objects in the generated image:
\begin{equation}
    \text{ObjectRecall} (G, I) = \frac{|V{\text{pred}} \cap V_{\text{gt}}|}{|V_{\text{gt}}|}, 
\end{equation}
where  $V_{\text{pred}}$  is the set of objects the LLM identifies as present in the image, and  $V_{\text{gt}}$  is the set of ground-truth objects from the original scene graph.

\noindent\textbf{Recall of Relationships.}
To further assess the quality of the generated scene, we evaluate the recall of relationships between objects in the image. 
For each relationship  $r \in E$, 
we check whether the predicted relationship  $r_{\text{pred}}$  exists between the corresponding objects in the generated image. 
The relationship recall is defined as:
\begin{equation}
    \text{RelationRecall} (G, I) = \frac{|E_{\text{pred}} \cap E_{\text{gt}}|}{|E_{\text{gt}}|}, 
\end{equation}
where  $E_{\text{pred}}$  represents the predicted relationships between objects in the generated scene, and  $E_{\text{gt}}$  represents the ground-truth relationships from the original scene graph. 
To obtain $E_{\text{pred}}$, we prompt the LLM with multiple-choice questions such as: \emph{``What is the relationship between the person and the sports ball in the image? A) kicking; B) throwing; C) holding; D) no visible relationship.''}

\noindent\textbf{SGScore.}
In addition to individual recalls of objects and relationships, we introduce a comprehensive metric, \emph{SGScore}, which evaluates the overall quality of the scene graph in terms of both objects and relationships. 
SGScore is computed as a weighted combination of object recall and relationship recall:
\begin{equation}
\begin{split}
    \text{SGScore} (G, I) &= \alpha \cdot \text{ObjectRecall} (G, I)  + \\ 
    &(1.0 - \alpha) \cdot \text{RelationRecall} (G, I), 
\end{split}
\end{equation}
where $\alpha \in [0, 1]$ is a hyperparameter that controls the relative importance of object recall versus relationship recall.
By adjusting this weight, we can tune the evaluation to place more emphasis on either the objects or the relationships, depending on the task requirements.
SGScore provides a holistic evaluation of how well the generated scene aligns with the scene graph, offering a balanced measure that reflects both object accuracy and relationship consistency.
\section{Scene Graph Feedback}
\label{sec:feedback}
Building on the scene graph-based evaluation, we propose a scene graph feedback to iteratively refine the generated image based on identified discrepancies between the image and the input scene graph.  
This process leverages multimodal LLMs to analyze the generated scene and provide targeted feedback for refinement. 

Specifically, given a scene graph $G=(V, E)$, 
we first perform scene composition using an LLM (see \cref{tab:prompt_sg_comp} in the \textbf{SMs}), 
in which nodes and edges are seamlessly integrated into a $prompt$ for an exact scene.
This $prompt$ results in an initial image $I_0$ through the diffusion model $f_D$.
With the input scene graph $G$ and the generated image $I_0$,
a multimodal LLM $f_M$ has been applied to evaluate the presence of objects and relationships.
The missing objects and relationships are constructed as a missing graph $G_{\text{miss}}$.
If discrepancies exists, \eg, $G_{\text{miss}} \neq (\emptyset, \emptyset)$, 
we will generate a reference image $I_1$ conditioned on the
$G_{\text{miss}}$ as does in generating $I_0$ conditioned on $G$.
To generate the final output image, we use IP-Adapter~\cite{DBLP:journals/corr/abs-2308-06721} to integrate $prompt$, $I_0$, and $I_1$, in which the cross attention process can be formulated as 
\begin{equation}
\begin{split}
    Z = &\mathrm{Attention}(Q, K_{prompt}, V_{prompt}) + \\
    &\lambda_0 \cdot \mathrm{Attention}(Q, K_{I_0}, V_{I_0}) + \\
    &\lambda_1 \cdot \mathrm{Attention}(Q, K_{I_1}, V_{I_1}), 
\end{split} 
\label{eq:feedback}
\end{equation}
where $Q$ is the query features of the latent variable, 
$K_{prompt}$ / $V_{prompt}$, $K_{I_0}$ / $V_{I_0}$,
$K_{I_1}$ / $V_{I_1}$,
are the projected features of $prompt$, $I_0$, $I_1$, respectively. 
The $\mathrm{Attention}$ is defined as 
$
\mathrm{Attention}(Q, K, V) = \mathrm{softmax}(\frac{Q K^T}{\sqrt{d}}) V
$ as does in \cite{DBLP:conf/nips/VaswaniSPUJGKP17}. 
$\lambda_0, \lambda_1$ are weight factors. 
\section{Experiments}
\subsection{Experimental Setup}
\textbf{Models.} 
We evaluate several popular diffusion models on Scene-Bench, including variants of Stable Diffusion and other state-of-the-art methods.

\noindent\textbf{Datasets.} 
We use the Visual Genome dataset~\cite{krishna2017visual}
following the data splits from SG2Im~\cite{johnson2018image}, and the proposed MegaSG dataset.
For a fair comparison, we balance samples for testing based on Scene Diversity and Scene Complexity (see \textbf{SMs}).

\noindent\textbf{Metrics.}
We employ common metrics such as Inception Score (IS)~\cite{salimans2016improved}, Fr\'{e}chet Inception Distance (FID)~\cite{heusel2017gans}, and CLIPScore~\cite{hessel2021clipscore}. 
Additionally, we introduce ObjectRecall, RelationRecall, and SGScore (see \cref{sec:evaluation} and \cref{sec:supp_exp} of the \textbf{SMs} for details).

\noindent\textbf{Scene Graph Representation.}
For text-to-image (T2I) models that condition on a sentence, we encode scene graphs in the format 
``\texttt{\{subject\} \{predicate\} \{object\}}''
(\eg, \texttt{cat sitting on desk, dog near chair}). 
For the prompt that converts a scene graph into a consistent description (\ie, scene composition in \cref{sec:feedback}), please refer to  \cref{sec:supp_exp} of the \textbf{SMs}.

\noindent\textbf{Scene Complexity.}
Based on Scene complexity defined as \cref{eq:comp}, we categorize the scene complexity into three levels: \emph{simple}, \emph{medium}, and \emph{hard} (details in \cref{sec:supp_exp} of the \textbf{SMs}).

\noindent\textbf{LLM.} We use Gemini 1.5 Flash~\cite{reid2024gemini} (cutoff November 2024) as the multimodal LLM in our experiments. We also report results using local multimodal LLMs like LLaVA~\cite{liu2024visual} in \cref{sec:supp_llm} of the \textbf{SMs}.
\subsection{Evaluation of Scene-Bench}
\begin{table*}[t]
\centering
\resizebox{0.9\textwidth}{!}{
\begin{tabular}{lcccccccc}
\toprule 
 \multirow{2}{*}{Method} & \multirow{2}{*}{Resolution}  & \multirow{2}{*}{IS $\uparrow$} & 
 \multirow{2}{*}{FID $\downarrow$} &    
 \multirow{2}{*}{CLIPScore $\uparrow$} &    \multicolumn{4}{c}{\textbf{SGScore} $\uparrow$}  \\
  \cline{6-9}
 & & & & & Overall & Simple (\# 3993) & Medium  (\# 930) & Hard (\# 173) \\ 
\midrule
SGDiff~\cite{yang2022diffusion} & 256x256 &   16.0 & 29.6&  - & 64.5 & 
64.2 & 66.3 & 61.5
\\
SceneGenie~\cite{farshad2023scenegenie} & 256x256 & 20.2 & 42.2 & - & - & - & - &-   \\ 
\hline 
  Composable~\cite{liu2022compositional} & 512x512 &20.5 & 47.5 & 22.0 & 48.0 & 48.9 &45.0 &44.5 \\ 
    Structured~\cite{feng2023training} & 512x512 &23.0 & 42.2 & 22.0& 52.5 &51.8 & 54.6 & 56.1 \\
  SD v1.5~\cite{rombach2022high} & 512x512 & 23.1 & 42.8 & 22.0   & 52.5 & 51.9 & 54.7 & 53.9 \\
  SD v2.1~\cite{rombach2022high} & 768x768 & 20.8 & 46.6&  22.1   & 54.4 & 53.5 & 57.8 & 57.9 \\
  PixArt-$\alpha$~\cite{chen2024pixart} & 1024x1024 & 20.8  &  52.9& 22.1   & 59.8 & 58.5 & 64.1 & 67.0 \\ 
  SD3.5~\cite{esser2024scaling} & 1024x1024 & 21.5  & 45.6 &  22.1 &     60.5 & 59.4 & 64.1 & 65.9 \\
    SD3~\cite{esser2024scaling} & 1024x1024 & 23.4 &44.5 & 22.1 &  62.1 & 60.9 & 66.3 & 66.7 \\
  SDXL~\cite{podellelbdmpr24} & 1024x1024 & 23.1 & 43.4& 22.1 &   60.7 & 59.6 & 64.0 & 69.6\\ \hline 
  RPG~\cite{yang2024mastering} (SDXL) & 1024x1024 & 22.9 & 44.2& 19.3
  &   69.3 ({+14.2\%})
  & 69.4 ({+16.4\%}) & 68.7 ({+7.3\%}) & 70.5 ({+1.3\%})\\ 
\rowcolor{gray!30}
  Ours (SD v1.5) & 512x512 & 20.7 & 41.6 &  19.1 &65.1 ({+24.0\%}) & 65.1 ({+25.4\%}) & 65.1 ({+19.0\%}) & 66.8 ({+23.9\%})\\ 
 \rowcolor{gray!30}
  Ours (SDXL) & 1024x1024 & 21.0  & 42.7 &  19.3 & 74.1 ({+22.1\%})
  & 74.2 ({+24.5\%})
  & 73.3 ({+14.5\%}) & 75.3 ({+8.2\%})\\ 
\bottomrule
\end{tabular}
}
\caption{Model Comparison on the VG test set. 
{Models including SGDiff and SceneGenie are trained on VG train set. 
Since SceneGenie~\cite{farshad2023scenegenie} does not release the code, we only present the reported IS and FID. 
}
}
\label{tab:benchmark_comparison_vg}
\end{table*}
\textbf{Performance on Visual Genome.}
Table~\ref{tab:benchmark_comparison_vg} presents the results on the VG test set. 
The first finding is that \emph{SGScore} provides much more distinguishability than other metrics like FID and CLIPScore. 
For instance, SD v1.5 has a better FID score than SD v2.1 (42.8 vs. 46.6), 
yet its SGScore is lower than that of SD v2.1 (52.5 vs. 54.4), 
indicating there are more missed objects and relationships in the images generated by SD v1.5. 
Another finding is that due to the VG test set being biased towards simple scenes, the performance on medium and hard scenes is counterintuitive: the performance should decrease as scene complexity increases, but this trend is not consistently observed, largely because of the limited number of complex scenes in the test set.
This counterexample justifies why we need a new large-scale benchmark to evaluate models comprehensively.

\noindent\textbf{Performance on MegaSG.}
\begin{table*}[t]
\centering
\resizebox{0.9\textwidth}{!}{
\begin{tabular}{lcccccccc}
\toprule 
 \multirow{2}{*}{Method} & \multirow{2}{*}{Resolution}  & \multirow{2}{*}{IS $\uparrow$} & 
 \multirow{2}{*}{FID $\downarrow$} & 
 \multirow{2}{*}{CLIPScore $\uparrow$} &    \multicolumn{4}{c}{\textbf{SGScore} $\uparrow$}  \\    \cline{6-9}
 & & & & & Overall & Simple (\# 15k) & Medium  (\# 20k) & Hard (\# 15k) \\ 
\midrule
  Composable~\cite{liu2022compositional} & 512x512 & 20.3 & 41.0 & 22.9
  &   42.0 & 61.0 & 39.0   & 28.3  
  \\ 
  Structured~\cite{feng2023training} & 512x512 &  28.6
  & 26.2 & 23.0 &   53.9 & 65.1 & 53.9  
  & 46.0  
  \\
  SD v1.5~\cite{rombach2022high} & 512x512 & 27.0  & 29.1  & 22.8    & 54.2 
  & 64.9 & 53.4   & 44.7  
  \\
  SD v2.1~\cite{rombach2022high} & 768x768 &  24.9 & 34.0 &  22.9  &   57.8
  & 68.2 & 56.4   & 49.2 
  \\
  PixArt-$\alpha$~\cite{chen2024pixart} & 1024x1024 & 24.3   &  43.9 &   23.0  &     59.5 
  & 68.4  & 58.2   & 52.7  
  \\  
  SD3.5~\cite{esser2024scaling} & 1024x1024 &  25.9 & 34.5 &  23.0 &   63.4 &
 73.1 & 61.9   & 55.7  
  \\
  SD3~\cite{esser2024scaling} & 1024x1024 &  27.2  &  35.5  & 23.0 &   65.2
  & 74.2 & 63.8  & 58.0   \\ 
SDXL~\cite{podellelbdmpr24} & 1024x1024 & 25.3  &31.6  & 23.0  &
65.6 & 72.9 & 64.6   & 59.6  \\  
  \hline 
RPG~\cite{yang2024mastering} (SDXL) & 1024x1024 & 23.4 &37.5 & 20.0 &
  71.0 (+8.2\%)
  & 76.5 (+4.9\%) &  70.5 (+9.1\%)&  66.1 (+10.9\%) \\ 
   \rowcolor{gray!30}
  Ours (SD v1.5) & 512x512 &  23.1  & 28.9 &  19.9
  & 62.0 ({+14.4\%})
  & 71.0 ({+9.4\%}) & 
  61.3 ({+14.8\%})  & 53.9 ({+20.6\%})
  \\     \rowcolor{gray!30}
  Ours (SDXL) & 1024x1024 & 21.6 & 34.1 &  20.0 &
  77.1 ({+17.5\%})
  &81.8 ({+12.2\%})
  &76.6 ({+18.6\%}) &73.1 ({+22.7\%}) \\
\bottomrule
\end{tabular}
}
\caption{Model comparison on a 50,000-image subset of the MegaSG dataset, sampled with a Scene Complexity of $\gamma=0$. Scene graph-based methods such as SGDiff~\cite{yang2022diffusion}, which are limited by the vocabulary of the VG dataset, were excluded from testing.}
\label{tab:benchmark_comparison_mega}
\end{table*}
To fairly assess models' abilities to handle more complex scenes, 
we evaluated them on a large-scale subset of the MegaSG dataset, 
sampled by Scene Complexity ($\gamma=0$). 
As shown in \cref{tab:benchmark_comparison_mega}, 
we observe a general decline in performance across all models. 
For example, SD v1.5's SGScore drops from 64.9 (simple scenes) to 44.7 (hard scenes), 
indicating they struggle more with accurately modeling the relationships in complex scenes. 
\begin{figure}[t]
    \centering
    \includegraphics[width=0.75\columnwidth]{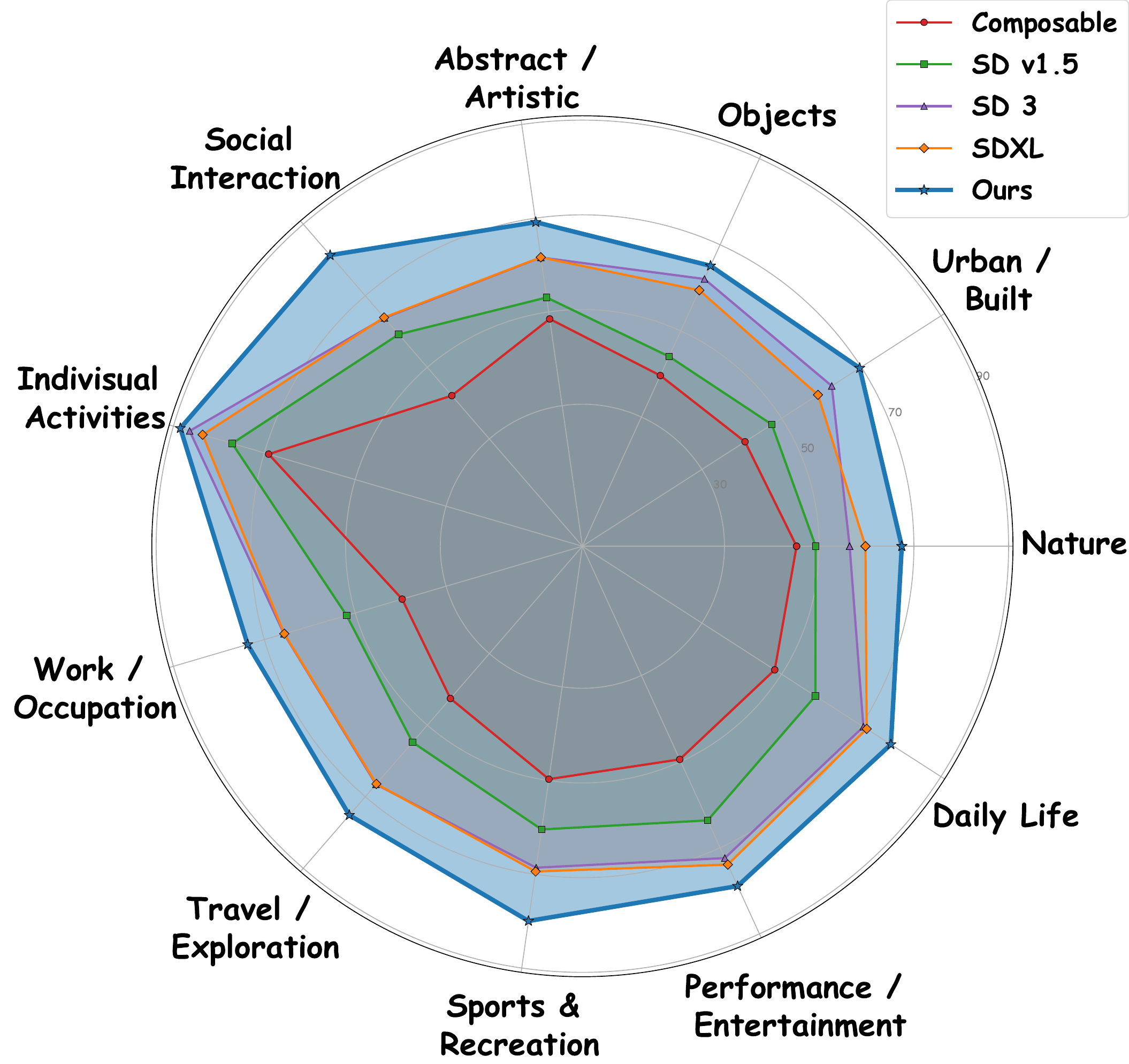}
    \caption{Comparison of model performances using SGScore across various scene categories.}
    \label{fig:radar-megasg}
\end{figure}

\noindent\textbf{Performance Across Scene Diversity.}
We evaluated model performance across diverse scene categories identified in our Scene Diversity analysis (see Section~\ref{sec:creation}). 
\cref{fig:radar-megasg} presents results for categories like \emph{Social Interaction}, \emph{Nature}, and \emph{Urban Environments}. 
Models generally perform better in categories like \emph{Individual Activities}, \emph{Performance / Entertainment}, and \emph{Daily Life}, 
but face challenges in \emph{Social Interaction} (where scenes often include multiple people) and \emph{Abstract / Artistic} (due to style discrepancies), \etc.
This variation underscores the importance of evaluating models across a broad range of scenarios to comprehensively assess their strengths and limitations.

\noindent\textbf{Impact of Scene Complexity.} 
Beyond the three coarse levels, we evaluate model performance across a complexity range from 1 to 10.
Detailed experimental results and analysis are provided in Section~\ref{sec:supp_eval_sgbench} of the \textbf{SMs}.
Our findings reveal that, although image quality (measured by FID) remains stable, increasing scene complexity significantly degrades the factual consistency of scene representations (measured by SGScore). 
Notably, our model consistently outperforms competitors by maintaining higher object and relationship recall across all complexity levels.
 


\begin{figure*}[t]
    \centering
    \definecolor{personcolor}{HTML}{FF6666}  
    \definecolor{objectcolor}{HTML}{66CCFF}  
    \newcommand{\imggap}{-12.5mm}
    \resizebox{0.965\textwidth}{!}
    {
    \begin{tikzpicture}
        \node (sg00) [align=center] at(0, 0) {
            \includegraphics[width=2.5cm]{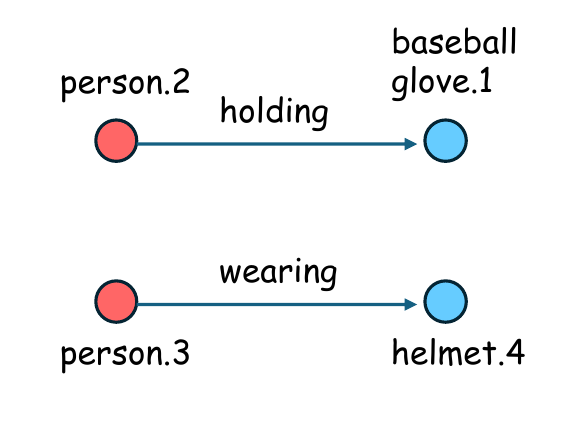}
        };
        \node (sg04) [align=center] at(0, -2.4) {
            \includegraphics[width=2.5cm]{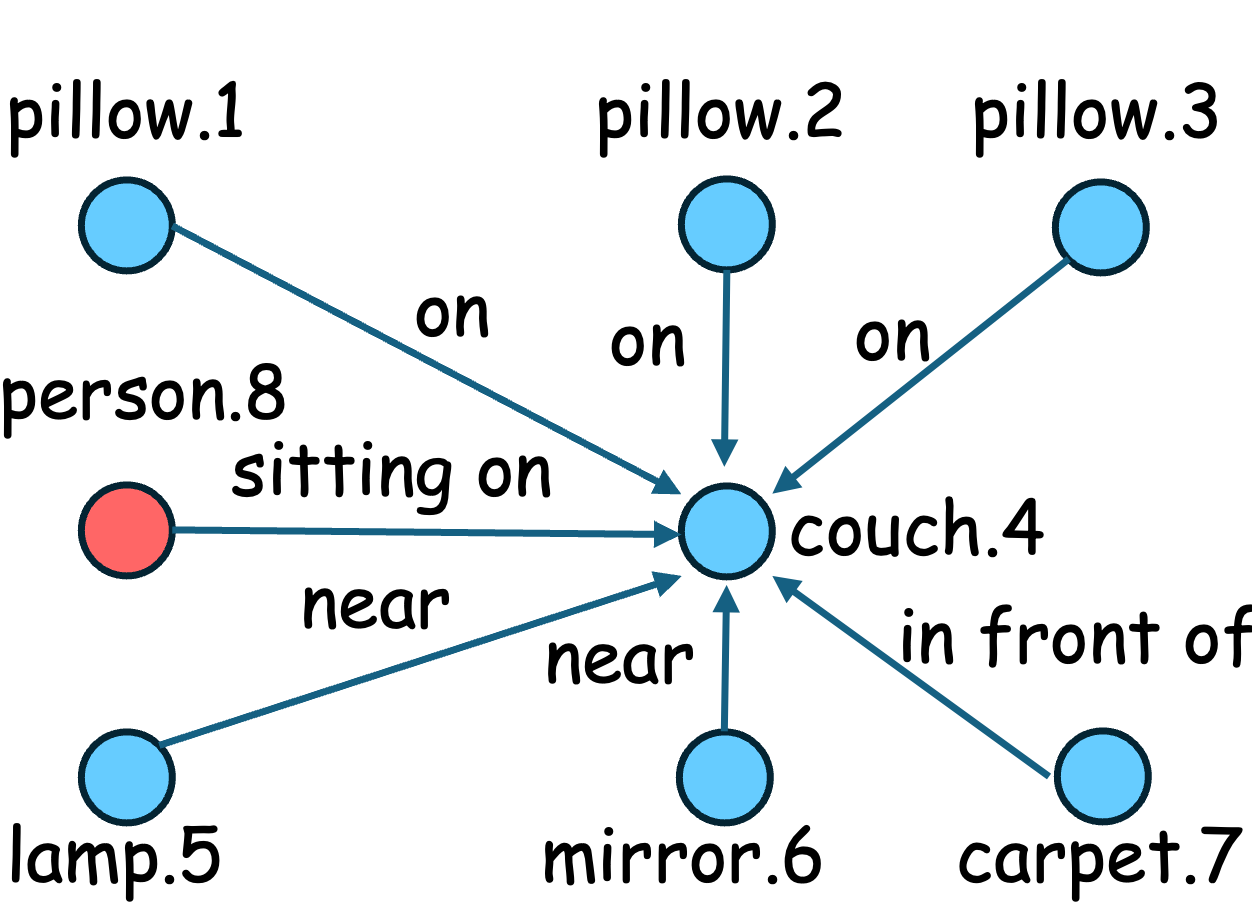}
        };     
        
        \node (n01) [align=center, right=of sg00, xshift=-12mm, yshift=0mm]  {
            \includegraphics[width=2cm, height=2cm]{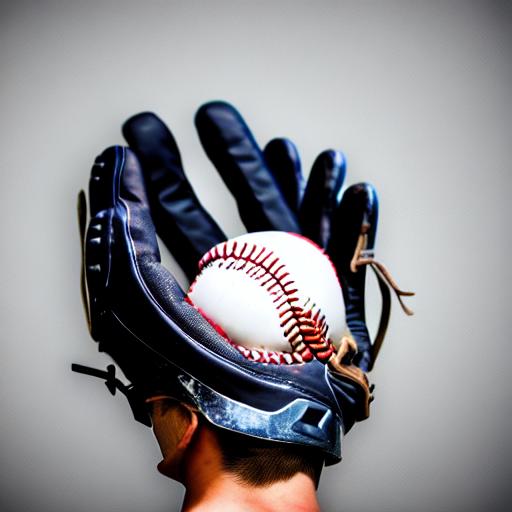}
        };
        \node [below=of n01, yshift=12mm]  {\small 62.5};
        
        \node (n02) [align=center, right=of n01, xshift=\imggap]     {
            \includegraphics[width=2cm, height=2cm]{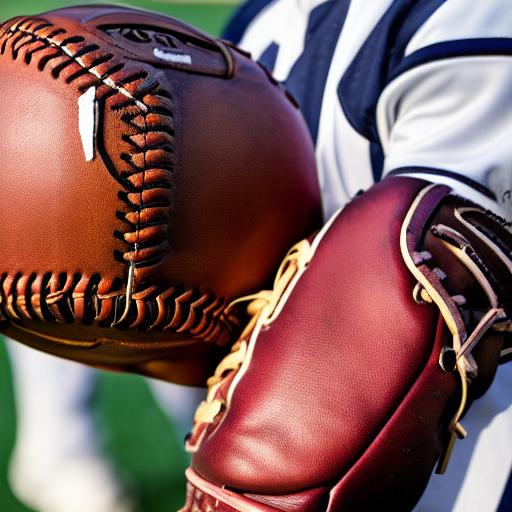}
        };
        \node [below=of n02, yshift=12mm]  {\small 50.0};
        
        \node (n03) [align=center, right=of n02, xshift=\imggap]  {
            \includegraphics[width=2cm, height=2cm]{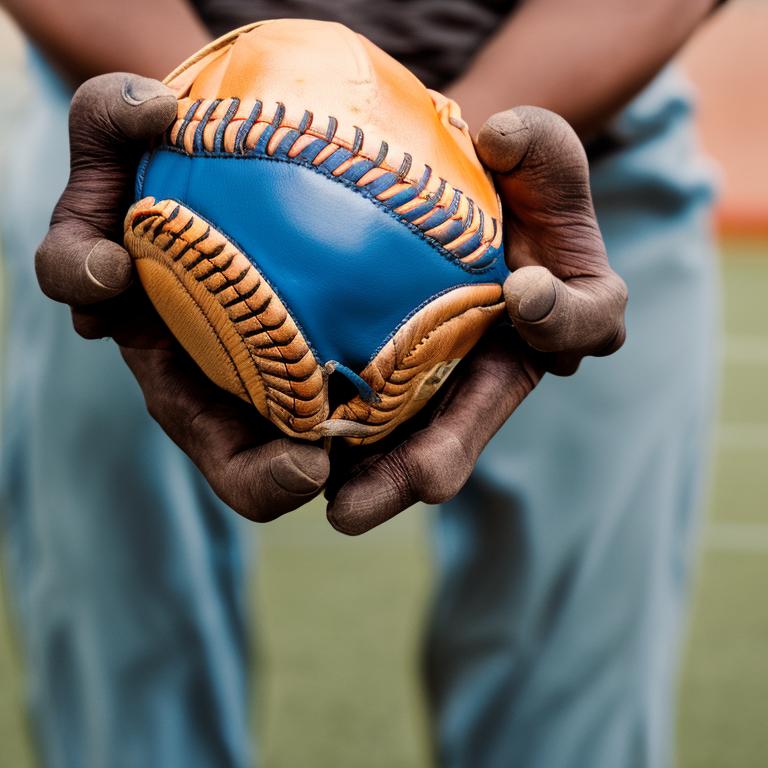} 
        };
        \node [below=of n03, yshift=12mm]  {\small 50.0};
        
        \node (n04) [align=center, right=of n03, xshift=\imggap]    {
            \includegraphics[width=2cm, height=2cm]{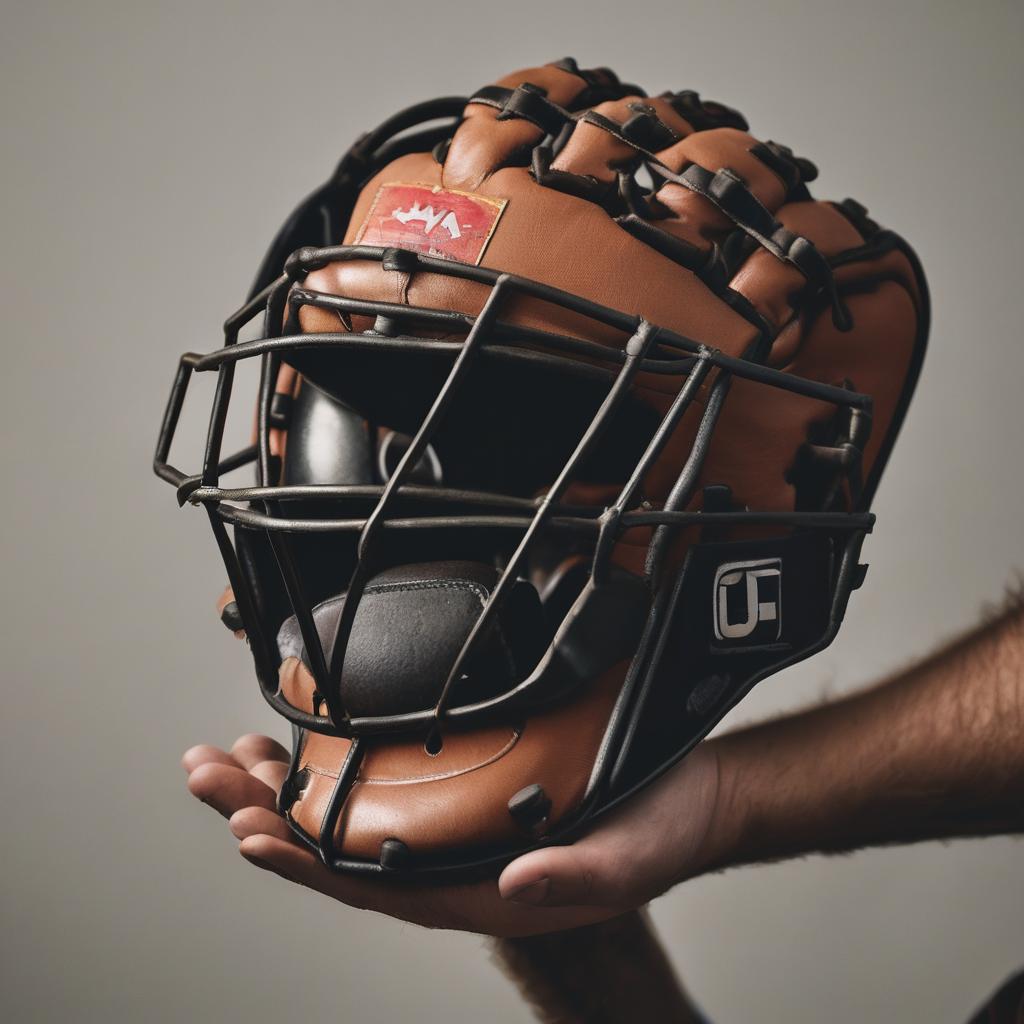} 
        };
        \node [below=of n04, yshift=12mm]  {\small 62.5};
        
        \node (n05) [align=center, right=of n04, xshift=\imggap]    {
            \includegraphics[width=2cm, height=2cm]{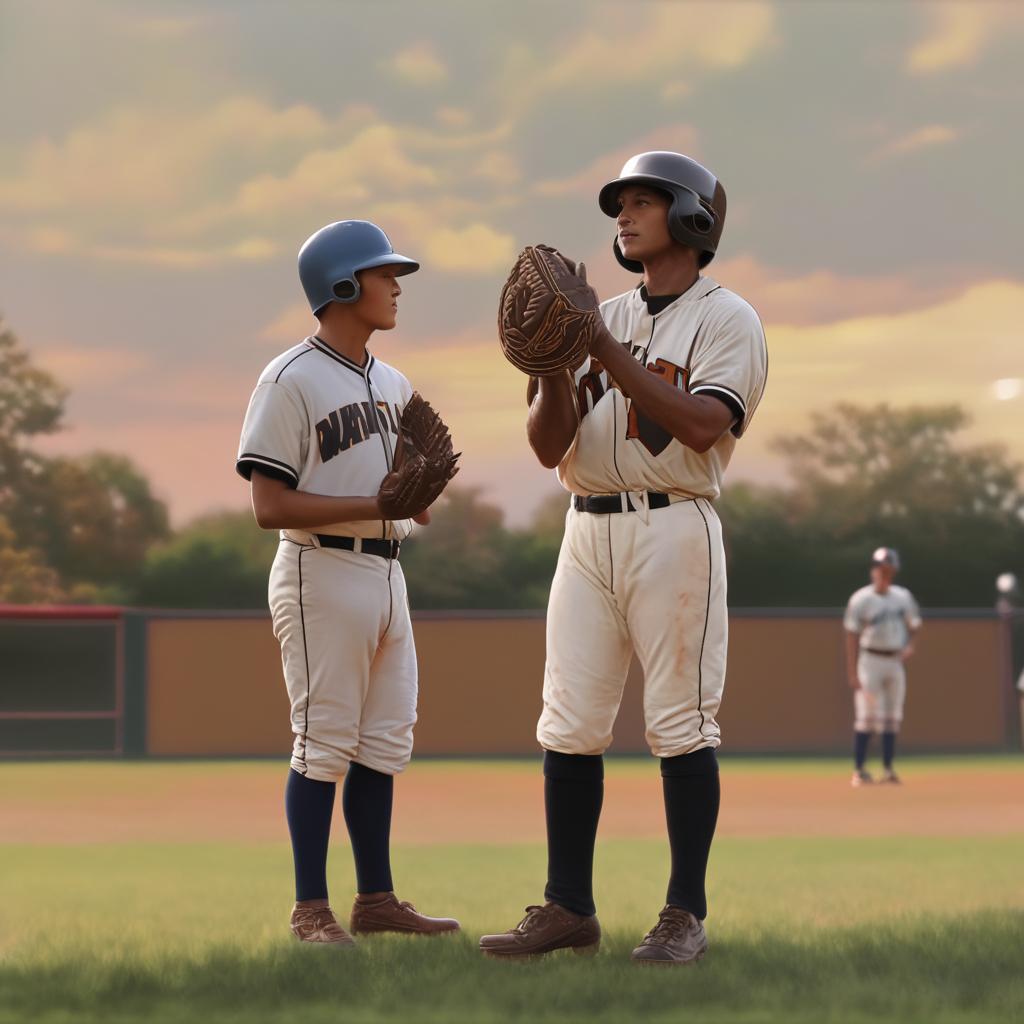} 
        };
        \node [below=of n05, yshift=12mm]  {\small 100.0};
        
        \node (t01) [above=of n01.north, yshift=-10.8mm] {Composable};
        \node [above=of n02.north, yshift=-10mm] {SD v1.5};
        \node [above=of n03.north, yshift=-10mm] {SD v2.1};
        \node [above=of n04.north, yshift=-10mm] {SDXL};
        \node [above=of n05.north, yshift=-10mm] {Ours};
        \node [left=of t01, xshift=5mm] {Scene Graph};

        \node (n41) [align=center, right=of sg04, xshift=-12mm, ]  {
            \includegraphics[width=2cm, height=2cm]{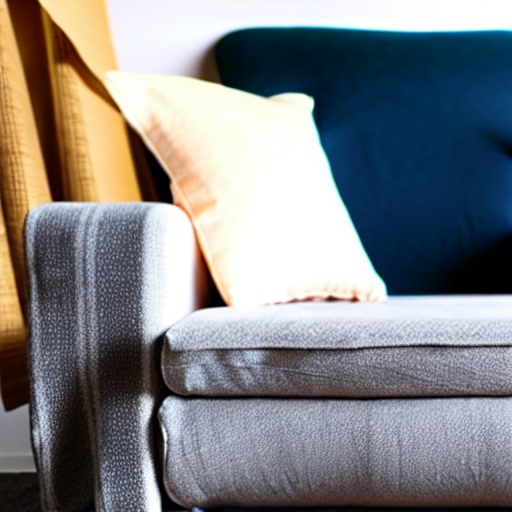}
        };
        \node [below=of n41, yshift=12mm]  {\small 19.6};
        
        \node (n42) [align=center, right=of n41, xshift=\imggap]     {
            \includegraphics[width=2cm, height=2cm]{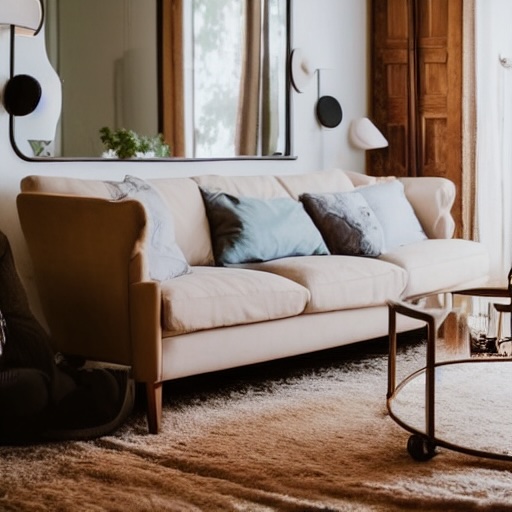}
        };
        \node [below=of n42, yshift=12mm]  {\small 66.0};
        
        \node (n43) [align=center, right=of n42, xshift=\imggap]  {
            \includegraphics[width=2cm, height=2cm]{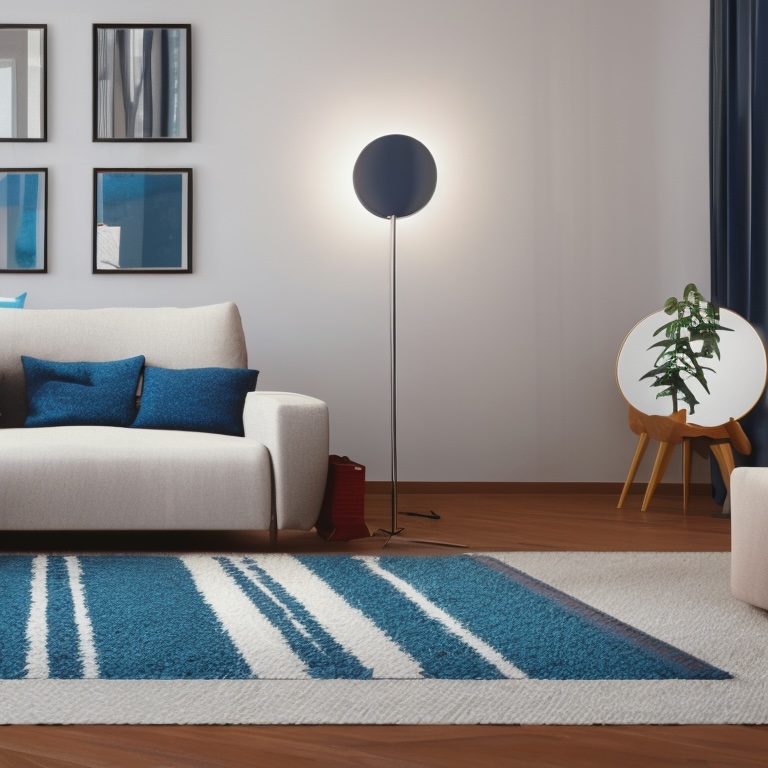}
        };
        \node [below=of n43, yshift=12mm]  {\small 73.2};
        
        \node (n44) [align=center, right=of n43, xshift=\imggap]    {
            \includegraphics[width=2cm, height=2cm]{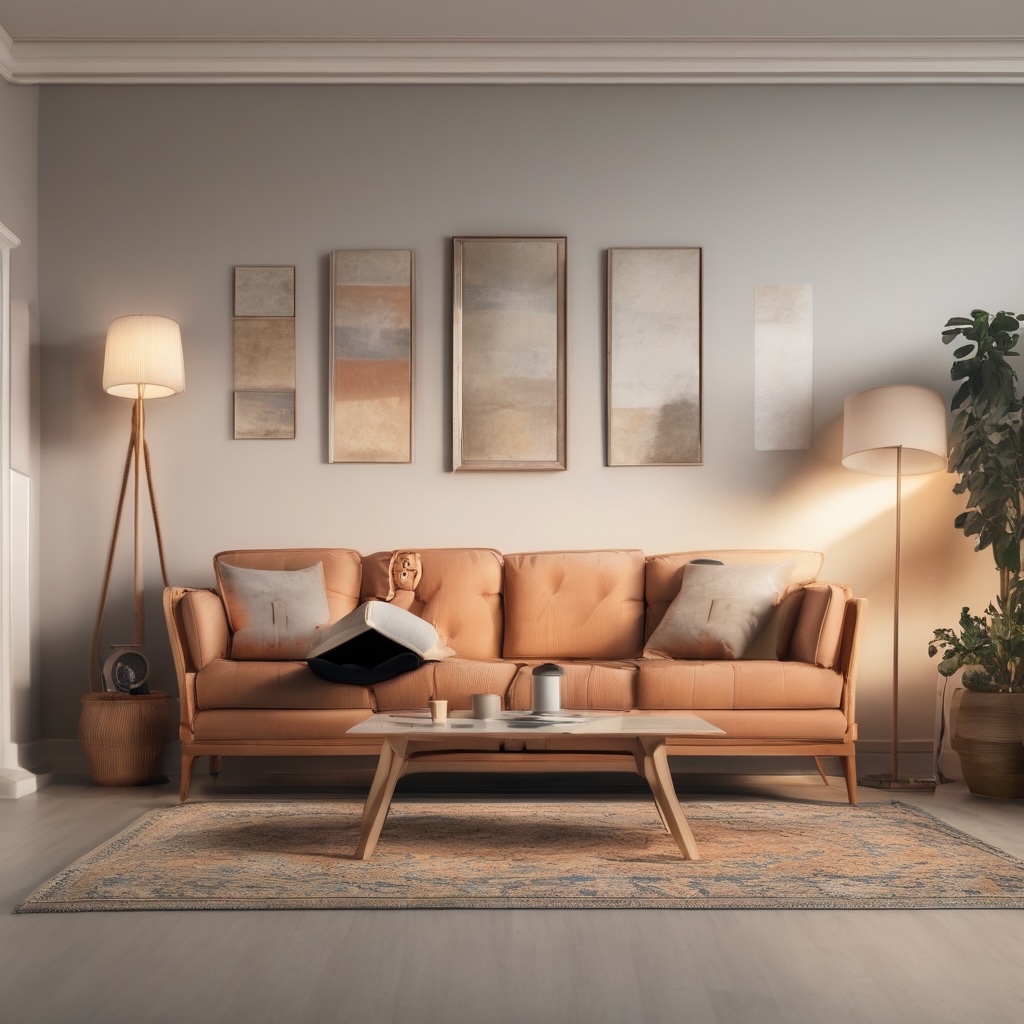}
        };
        \node [below=of n44, yshift=12mm]  {\small 59.8};
        
        \node (n45) [align=center, right=of n44, xshift=\imggap]    {
            \includegraphics[width=2cm, height=2cm]{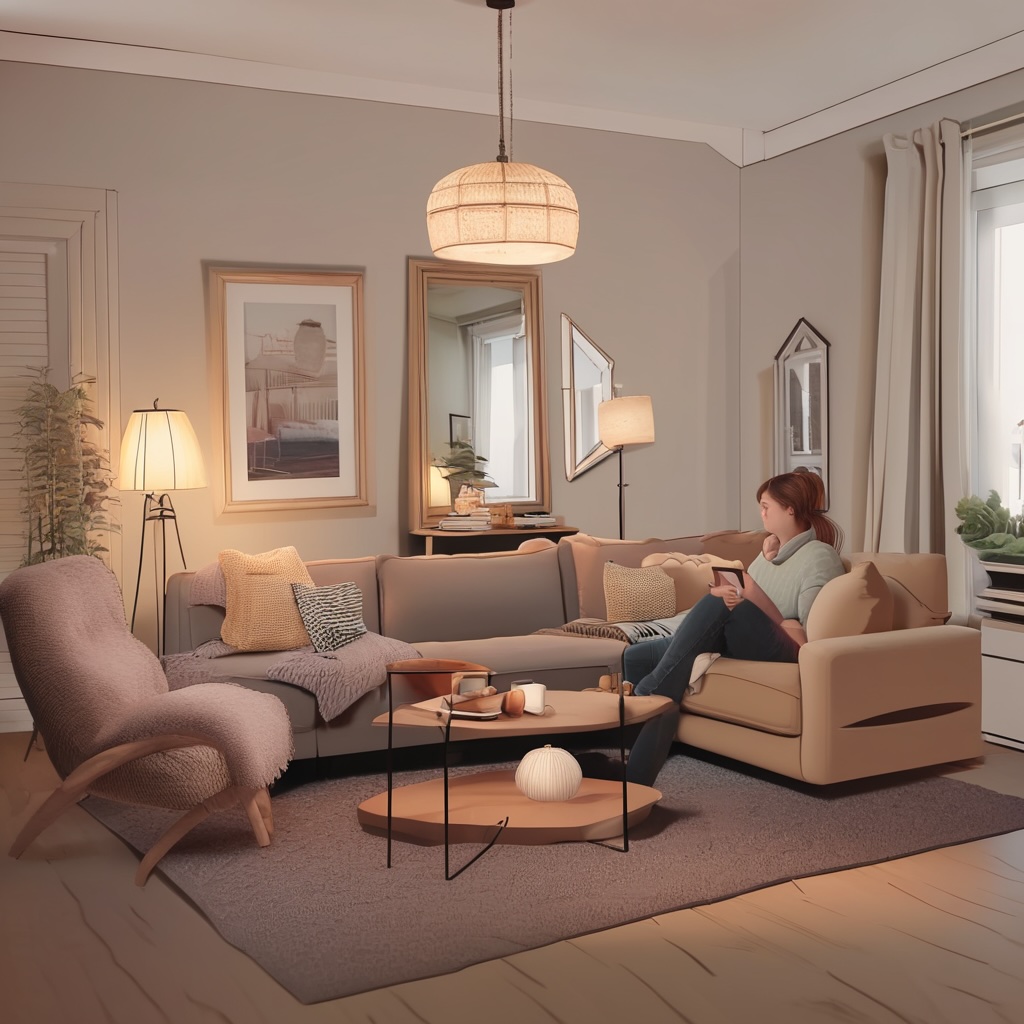}
        };
        \node [below=of n45, yshift=12mm]  {\small 100.0};    
    \end{tikzpicture}
    }
\caption{Comparison of Scene Graph-based Image Generation across Different Models. 
Each row displays a unique scene graph used as input for image generation. We present the \textbf{SGScore} below each generated image to quantify the consistency between the scene graph and the generated output. 
}
    \label{fig:qualitative_results}
\end{figure*}
\subsection{Evaluation of Scene Graph Feedback}
We evaluated our scene graph feedback pipeline using two diffusion models: SD v1.5~\cite{rombach2022high} and SDXL~\cite{podellelbdmpr24}. For each model, we compared three settings: baseline (without scene composition or feedback), with scene composition only, and with both scene composition and feedback.

\noindent\textbf{Results and Analysis.} Table~\ref{tab:feedback_exp} summarizes the results. For SD v1.5, the baseline achieved an ObjectRecall of $64.93\%$ and a RelationRecall of $44.19\%$ (SGScore $54.56\%$). Incorporating scene composition improved these metrics to $75.45\%$ and $48.84\%$ (SGScore $62.14\%$), demonstrating that detailed prompts help the model better capture specified objects and relationships. Applying our feedback strategy further increased ObjectRecall to $79.93\%$ and RelationRecall to $53.97\%$ (SGScore $66.95\%$), indicating effective correction of discrepancies.
A similar trend was observed in SDXL, where improvements after applying scene composition and feedback increased the SGScore from $65.50\%$ to $77.25\%$.

Compared with LLM-based methods like RPG~\cite{yang2024mastering}, the performance gain on VG or MegaSG is significant.  
RPG~\cite{yang2024mastering} utilizes an LLM as an agent to perform re-captioning, region planning and merging, while it lacks a feedback for ensuring factual consistency.

These results demonstrate that our scene graph feedback  effectively enhances factual consistency by identifying and correcting discrepancies between generated images and the intended scene graphs.

\noindent\textbf{Ablation Study of IP-Adapter.} 
To assess the impact of the additional parameters introduced by the IP-Adapter, 
we conduct an ablation study. 
Detailed results are provided in \cref{sec:ablation_ip} of the \textbf{SMs}.
\begin{table}[t]
\centering
\resizebox{\columnwidth}{!}{
\begin{tabular}{lccc}
\toprule 
{Model} & {ObjectRecall} $\uparrow$ & {RelationRecall} $\uparrow$ & SGScore $\uparrow$ \\ 
\midrule
\multicolumn{4}{l}{{SD v1.5~\cite{rombach2022high}}} \\
\quad Baseline & $64.93\pm0.31$ & $44.19\pm0.09$ & $54.56\pm0.12$ \\
\quad + Scene Composition & $75.45\pm0.19$ & $48.84\pm0.39$ & $62.14\pm0.25$ \\ 
   \rowcolor{gray!30}
\quad + Feedback & $\mathbf{79.93}\pm0.34$ & $\mathbf{53.97}\pm0.20$ & $\mathbf{66.95}\pm0.23$ \\
\midrule
\multicolumn{4}{l}{{SDXL~\cite{podellelbdmpr24}}} \\
\quad Baseline & $77.22\pm0.22$ & $53.78\pm0.36$ & $65.50\pm0.19$ \\
\quad + Scene Composition & $88.07\pm0.17$ & $60.37\pm0.14$ & $74.22\pm0.14$ \\
   \rowcolor{gray!30}
\quad + Feedback & $\mathbf{91.30}\pm0.24$ & $\mathbf{63.20}\pm0.10$ & $\mathbf{77.25}\pm0.21$ \\
\bottomrule
\end{tabular}
}
\caption{Effectiveness of the scene graph feedback on 5,000 images sampled from MegaSG.}
\label{tab:feedback_exp}
\end{table}
\subsection{Qualitative Evaluation}
We present qualitative results to demonstrate the effectiveness of Scene-Bench and the proposed scene graph feedback. 
Fig.~\ref{fig:qualitative_results} compares images generated by various models using the same scene graphs. 
Most of models often struggle with complex scenes, leading to images with missing objects or incorrectly depicted relationships. 
For example, when generating a scene from the scene graph \textit{\(<\)person.2, holding, baseball\ glove.1\(>\), \(<\)person.3, wearing, helmet.4\(>\)}, 
previous models may omit the \textit{helmet} or fail to represent \textit{person wearing helmet}.

With the proposed scene graph feedback, the generated image more faithfully represents the intended scene graph. 
The feedback process identifies missing elements and corrects relational inaccuracies, resulting in the image where one person is wearing a helmet and another is holding a baseball glove.
This demonstrates the model's improved ability to handle complex object interactions and spatial arrangements, highlighting the benefits of our approach.
\subsection{Human Evaluation}
To validate the efficacy of \emph{SGScore} in improving the verification of factual consistency and assess the impact of the proposed feedback pipeline, we conducted a human evaluation. Annotators were presented with 1,000 four-to-one comparative queries (see examples in \cref{sec:supp_human} of the \textbf{SMs}). Each query displayed an original image alongside four generated images produced by different models. Annotators were instructed to select the image that most accurately preserved the object presence and relationships of the original. 
As illustrated in \cref{fig:confusion} of the \textbf{SMs}, both human judgments and machine-based selections consistently favored our model, thereby confirming its superior factual consistency as measured by \emph{SGScore}.
\section{Discussion}
In the field of SG2IM, most existing approaches~\cite{johnson2018image,ashual2019specifying,yang2022diffusion,farshad2023scenegenie,shen2024sg,liu2024r3cd,wang2024scene} focus primarily on network architecture design, train on widely used datasets (\eg, Visual Genome and COCO-stuff), and report conventional metrics such as IS, FID, and CLIPScore.
However, \emph{assessing factual consistency and leveraging  identified inconsistency as feedback remain underexplored}. 
To address this gap, we introduce a large-scale scene graph dataset, a novel metric \emph{SGScore}, and an automatic evaluation pipeline to systematically measure the factual consistency in SG2IM.
Based on the evaluation pipeline, we incorporate a training-free feedback pipeline to enhance factual consistency.

One potential concern for the newly introduced dataset, \emph{MegaSG}, is its lack of attribute annotations (\eg, color, shape, texture) that are present in the VG dataset.
To assess the impact of this omission, we present an experimental analysis in \cref{sec:supp_attr} of the \textbf{SMs}.
The results indicate that explicit attribute binding yields a marginal difference.
Considering the trade-off between additional annotation costs and minimal gains, 
we opted to omit attributes in the construction of \emph{MegaSG}.
Beyond this concern, we compare \emph{Scene-Bench} with existing T2I benchmarks like TIFA~\cite{hu2023tifa} and DSG~\cite{cho2024dsg} in \cref{sec:supp_benchmarks} 
to validate the reliability of \emph{SGScore}.
\section{Conclusion}
In this work, we introduce \emph{Scene-Bench}, 
a comprehensive benchmark for evaluating the factual consistency of generating natural scenes from scene graphs. 
Our benchmark incorporates a large-scale dataset, \emph{MegaSG}, 
with a novel metric, \emph{SGScore}, 
which quantitatively assesses both the presence of objects and the accuracy of relationships through the reasoning capabilities of multimodal LLMs. 
Building upon this evaluation, our scene graph feedback mechanism iteratively refines generated images by correcting inconsistencies between the scene graph and the output. 
This process significantly improves the factual consistency. 
Extensive experiments demonstrate that \emph{Scene-Bench} offers a rigorous evaluation framework, especially in complex scenes where traditional metrics fall short.
We believe that \emph{Scene-Bench} will establish a new standard and inspire future research in high-fidelity, controllable generation. 
{
    \small
    \bibliographystyle{ieeenat_fullname}
    \bibliography{ref}
}

\clearpage
\setcounter{page}{1}
\maketitlesupplementary

\appendix
%

\begin{figure*}[t]
    \centering
    \includegraphics[width=\textwidth]{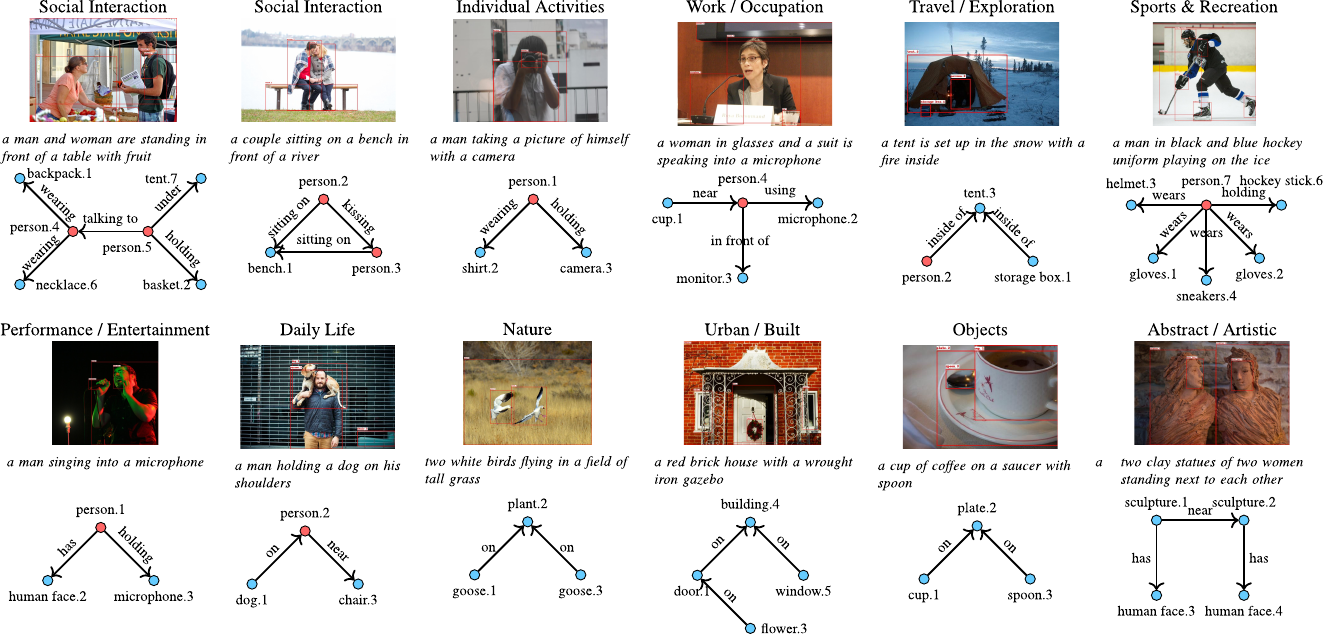}
    \caption{Illustration of scene categories in the MegaSG dataset. The image shows various themes, such as People-Centric (\eg, social interaction, individual activities) and Non-People-Centric (\eg, nature, urban environments). The caption is provided for illustrative purposes and generated using BLIP-2~\cite{li2023blip}, and the scene graph is constructed as described in  \cref{sec:megasg} and \cref{sec:supp_mega}.
    }
    \label{fig:scene_cat}
\end{figure*}
\begin{figure*}[t]
    \centering
    \begin{subfigure}[t]{0.48\textwidth}
        \centering
        \includegraphics[width=\textwidth]{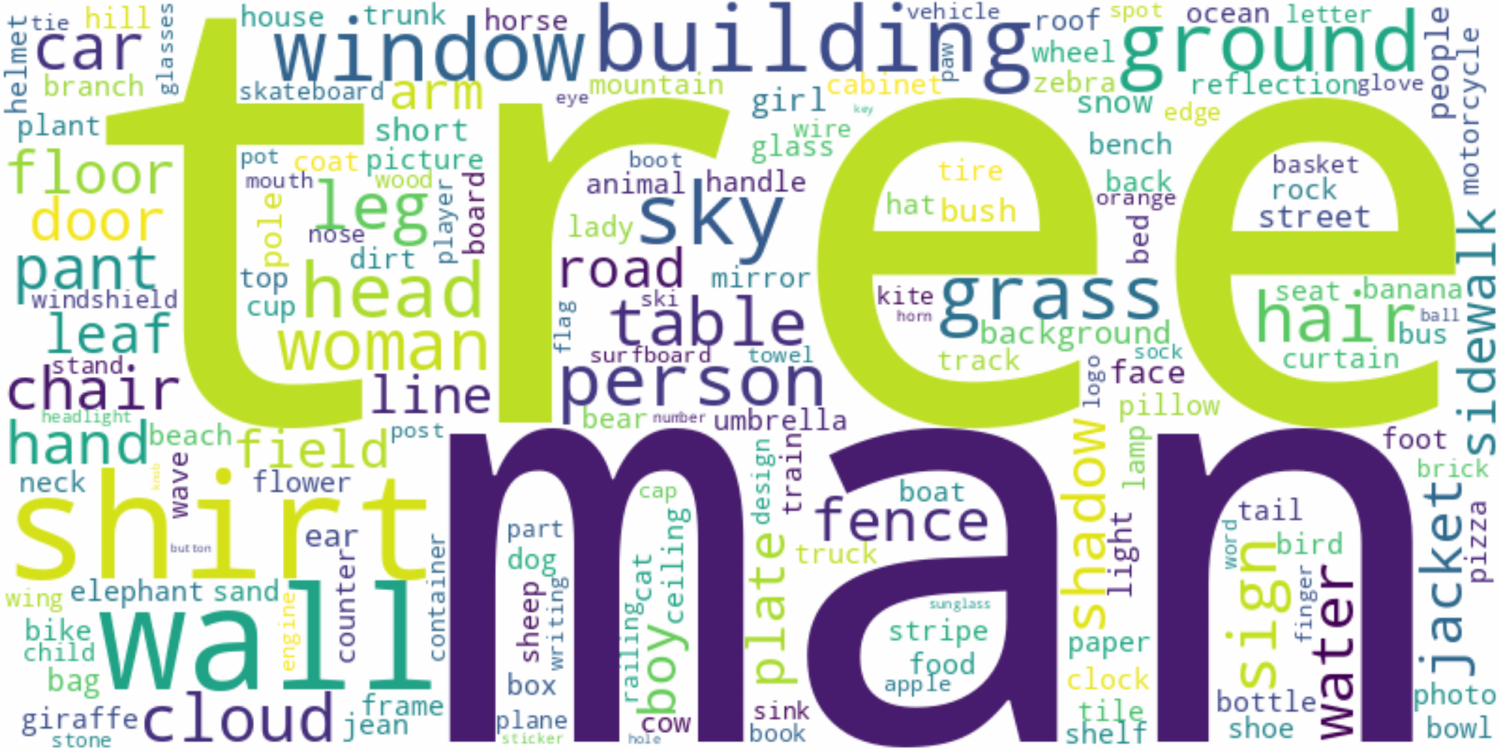}
        \caption{VG Objects}
        \label{fig:vg_obj}
    \end{subfigure}
    \hfill
    \begin{subfigure}[t]{0.48\textwidth}
        \centering
        \includegraphics[width=\textwidth]{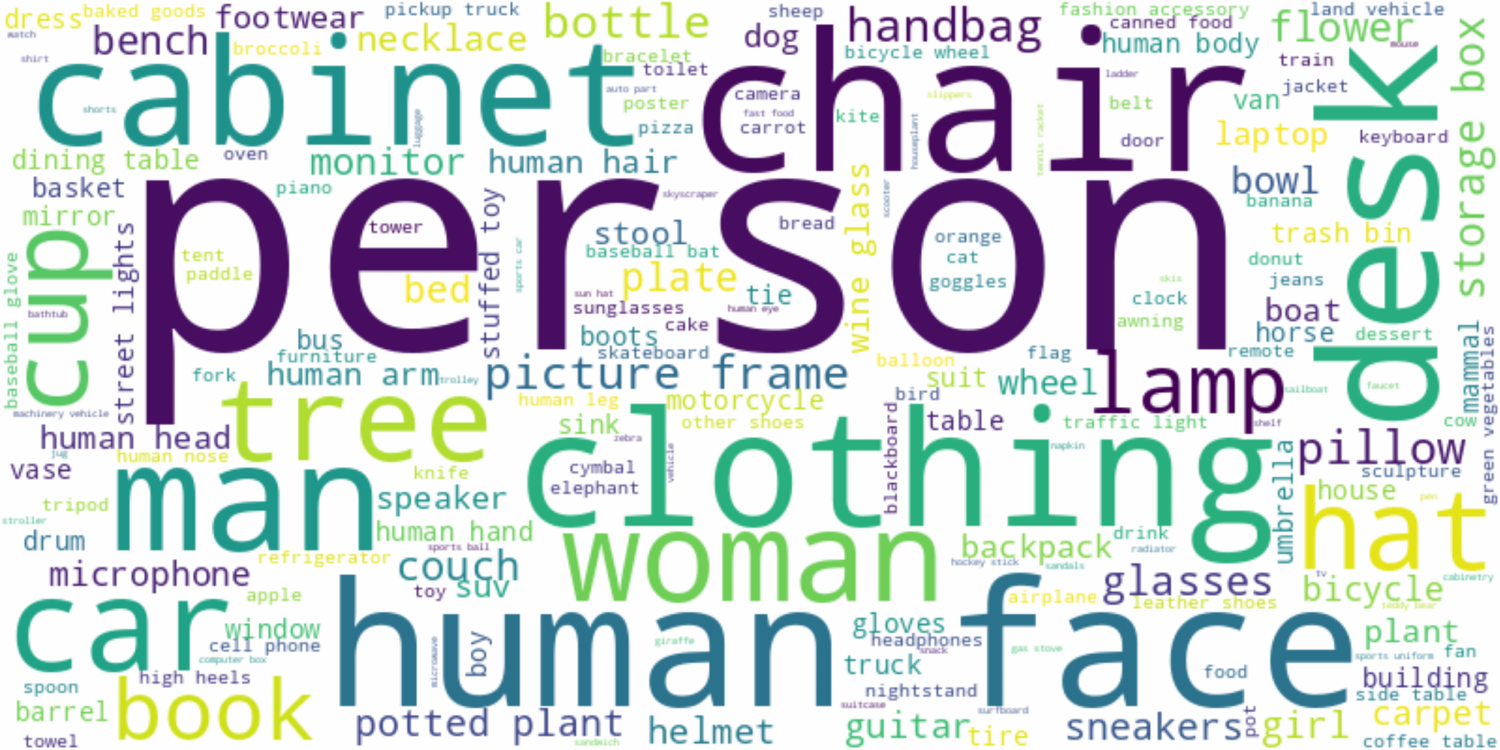}
        \caption{MegaSG Objects}
        \label{fig:mega_obj}
    \end{subfigure}
    
    \vspace{1em} 

    \begin{subfigure}[t]{0.48\textwidth}
        \centering
        \includegraphics[width=\textwidth]{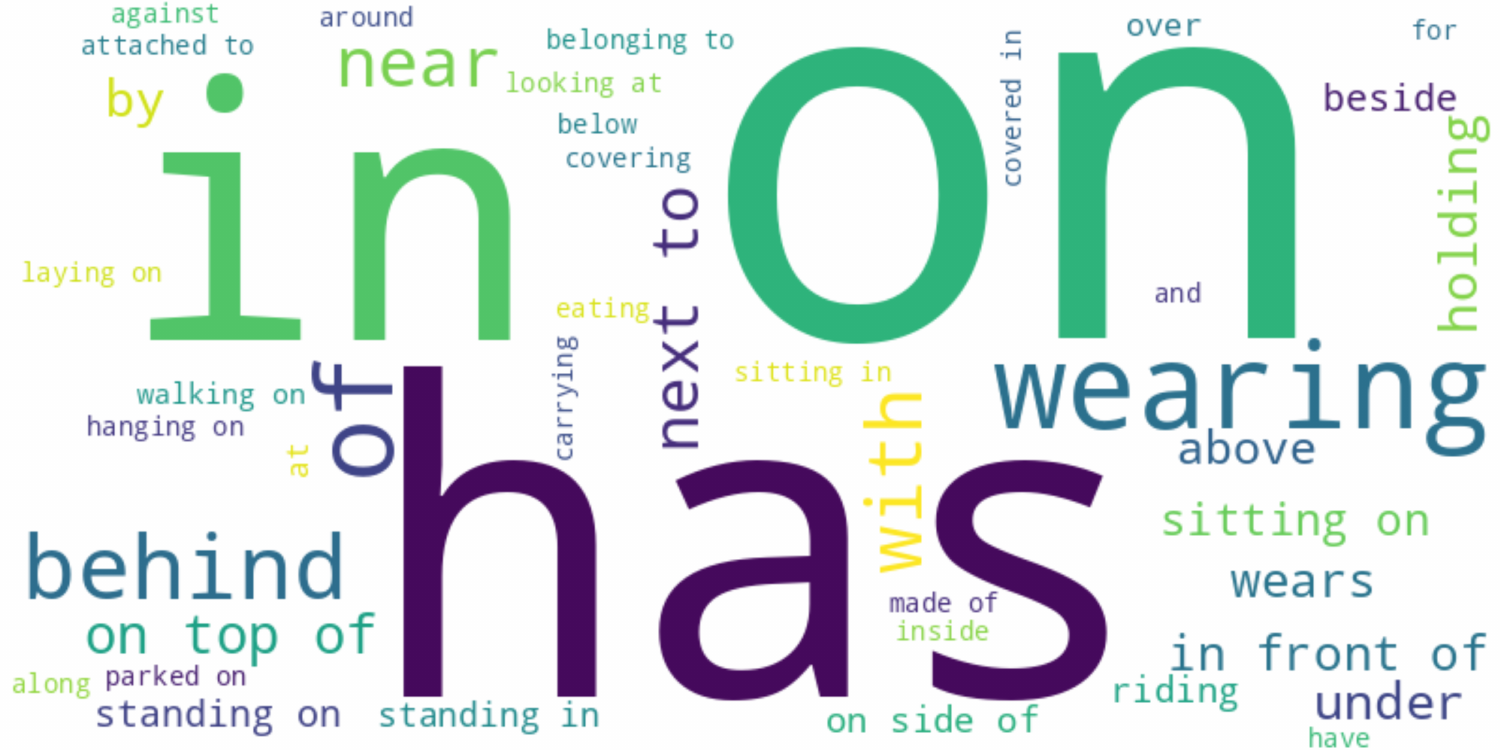}
        \caption{VG Relationships}
        \label{fig:vg_rels}
    \end{subfigure}
    \hfill
    \begin{subfigure}[t]{0.48\textwidth}
        \centering
        \includegraphics[width=\textwidth]{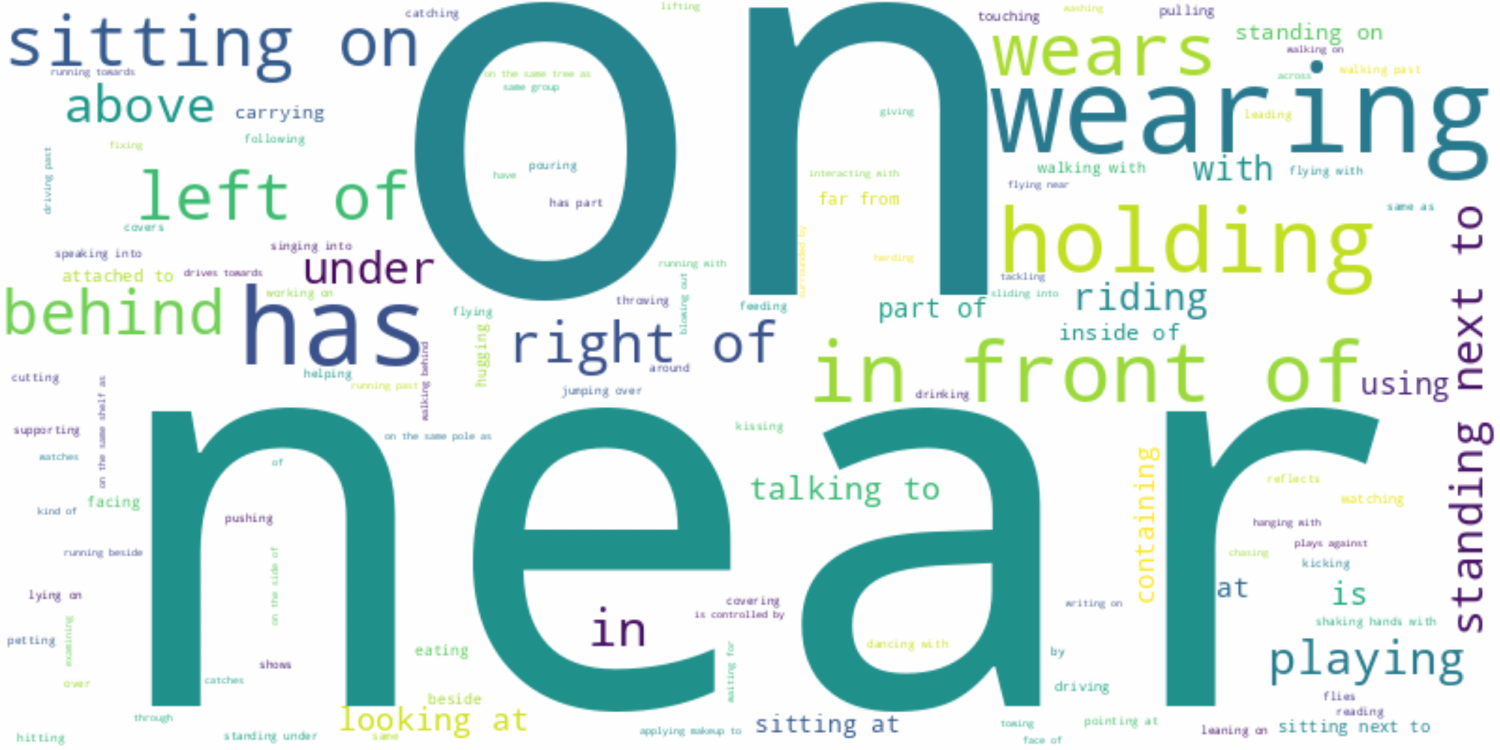}
        \caption{MegaSG Relationships}
        \label{fig:mega_rels}
    \end{subfigure}
    
    \caption{Word clouds of objects and relationships in the Visual Genome (VG) and MegaSG datasets.
    (a) and (b) illustrate the diversity of objects, while (c) and (d) highlight the relationships. The comparison demonstrates MegaSG's broader vocabulary and richer representation of object-relationship semantics.
    }
    \label{fig:wordcloud}
\end{figure*}
\section{MegaSG: a large-scale dataset of scene graphs}
\label{app:mega}
\label{sec:supp_mega}
\textbf{Creation of the Dataset.}
\begin{algorithm}[t]
\caption{Generate Scene Graph}
\label{alg:generate_scene_graph}
\begin{lstlisting}[style=PythonCommentStyle]
import google.generativeai as genai

generation_config = {
    "temperature": 0.7, "top_p": 0.95,
    "top_k": 64, "max_output_tokens": 8192,
    "response_mime_type": "application/json",
}
# Load the generative model
model = genai.GenerativeModel(
        "gemini-1.5-flash", 
        generation_config=generation_config
)

prompt_template = """Given a set of detected objects in an image, each object is characterized by a name, a bounding box in "(xmin, ymin, xmax, ymax)" format. Please generate a scene graph to describe this image. The scene graph should describe relationships in the format "source -> relation -> target". Example Output:\n{"relationships": [{"source": "object_id1", "target": "object_id2", "relation":\n"relation_type"}, ... ]}\n Now, objects are {OBJECTS}. The original width and height of the provided image are {IMG_WH}. Please output the scene graph in JSON style without any comments."""

def annotate(image_name):
    """
    image_name: file path of the image
    """
    # Load image and get its dimensions
    image = Image.open(image_name)
    image_wh = (image.width, image.height)

    # Load objects in the image,   
    # e.g., {["sports ball.1:[312, 360, 370, 417]", "person.2:[116, 49, 309, 491]", "person.3:[367, 108, 550, 477]"]}
    image_objects = load_objects(image_name)

    # Construct text prompt 
    text_prompt = prompt_template.replace(
        "OBJECTS", str(image_objects)).replace(
        "IMG_WH", str(image_wh))

    # Generate scene graph using generative model
    response = model.generate([
        image, text_prompt])
    return response
\end{lstlisting}
\end{algorithm}
We construct the \textbf{MegaSG} dataset by leveraging three widely used object detection datasets: COCO~\cite{chen2015microsoft}, Object365~\cite{shao2019objects365}, and Open Images v6~\cite{kuznetsova2020open}. These resources offer diverse scenes with meticulously annotated objects. To ensure the reliability of the scene graphs, we discard images containing fewer than three objects. The annotation prompt used to generate scene graphs from images is detailed in \cref{alg:generate_scene_graph}.

Specifically, the multimodal large language model receives an image—along with associated object categories and bounding boxes—as input and produces a scene graph that depicts the relationships between objects. 
For example, as illustrated in \cref{fig:scene-graph-example}, 
the model identifies a ``person" kicking a ``sports ball" and another ``person" nearby, 
yielding the scene graph:
\emph{\{``source": ``person.2", ``target": ``sports ball.1", ``relation": ``kicking"\}, \{``source": ``person.2", ``target": ``person.3", ``relation": ``near"\}}. 
This structured representation captures both the objects and their spatial as well as relational interactions.

\noindent\textbf{Scene Diversity.}
To classify the scene categories, we utilize an LLM (\eg, Gemini 1.5 Flash~\cite{reid2024gemini}) to perform such classification.
The prompt used here is 
\emph{
Now, we have a list of image information like \{IMAGE\_INFO\} , where each image information contains ``xyxy'' bounding boxes and ``relationships'' depicting the relation between the ``source'' object and the ``target'' object. Please classify the scene in **each image** using the following hierarchy:
Level 1:
- People-Centric,
- Non-People Centric.
Level 2:
If People-Centric: [Choose one: Social Interaction, Individual Activities, Work/Occupation, Travel/Exploration, Sports \& Recreation, Performance/Entertainment, Daily Life];
If Non-People Centric: [Choose one: Nature, Urban/Built, Objects, Abstract/Artistic].
Please provide the classification for each image in the list, and present your answer as a **JSON-formatted** list of dictionaries, where each dictionary corresponds to an image and contains the following keys: ``image\_id'', ``file\_name'', ``level 1'', ``level 2''.
}

\cref{fig:scene_cat} illustrates examples of categorized scenes in MegaSG, 
showcasing the diversity and range of scenarios covered in the dataset.
\begin{table*}[t]
    \centering
   \resizebox{\textwidth}{!}
    {
    \begin{tabular}{l|c|ccc|ccc|ccc|ccc}
    \toprule
         \multirow{2}{*}{SGG model}&  \multirow{2}{*}{Training Data} &  
         \multicolumn{6}{c|}{SGDet} 
         & \multicolumn{6}{c}{PredCls}
         \\ 
         &  &  
         \multicolumn{3}{c|}{R@20/50/100} & 
         \multicolumn{3}{c|}{mR@20/50/100} &
         \multicolumn{3}{c|}{R@20/50/100} & 
         \multicolumn{3}{c}{mR@20/50/100} 
         \\
    \midrule
         LSWS  \cite{yelinguistic}&   & - & 3.28 & 3.69   
         &    \multicolumn{3}{c|}{-} &
          \multicolumn{3}{c|}{-} &  \multicolumn{3}{c}{-}  
         \\ 
         MOTIFS \cite{zellers2018neural}&       & 5.02 & 6.40 & 7.33   &  \multicolumn{3}{c|}{-}&
          \multicolumn{3}{c|}{-} &  \multicolumn{3}{c}{-}  
         \\ 
         Uniter \cite{chen2020uniter} &COCO \cite{chen2015microsoft}
              & 5.42 & 6.74 & 7.62  &   \multicolumn{3}{c|}{-}&\multicolumn{3}{c|}{-} &  \multicolumn{3}{c}{-}  \\ 
         $\text{VS}^3_{\text{(Swin-T)}}$  \cite{zhang2023learning}   &  Caption   & 4.56 & 5.79 & 6.79  & 2.18 & 2.59 & 3.00 & 12.30 & 16.77
         & 19.40 & 3.56 & 4.79 & 5.51 
         \\ 
         $\text{VS}^3_{\text{(Swin-L)}}$  \cite{zhang2023learning}  &  (104k)     & 4.82 & 6.20 & 7.48  & 2.29  &  2.70  &  3.09
         & 12.54 & 17.28 & 19.89 
         & 3.57 & 4.83 & 5.56 
         \\ 
         $\text{OvSGTR}_\text{(Swin-T)}$ \cite{chen2024expanding}  &    &  {{6.61}} & {{8.92}} & {{10.90}}  & 1.09   &  1.53  & 1.95 
         & 16.65 & 22.44 & 26.64 & 2.47 & 3.58 & 4.41 
         \\ 
        $\text{OvSGTR}_\text{(Swin-B)}$ \cite{chen2024expanding} &    &   {6.85}
          &    {9.33}  &  {11.47}   
          &   {1.28}   &   {1.79}
          &  {2.18}  &
          16.82 & 22.79 & 27.04 & 2.94 & 4.24 & 5.26  
         \\ 
      \hline 
        $\text{VS}^3_{\text{(Swin-T)}}$  \cite{zhang2023learning}    &  & 5.56
             &  8.19      &   10.17 &  1.15  &   1.71  & 2.20  & 
             23.81 & 29.64 & 32.18 
             & 4.70 & 5.96 & 6.57 
             \\ 
        $\text{VS}^3_{\text{(Swin-L)}}$  \cite{zhang2023learning}    & {MegaSG} & 9.74
             &  14.80      & 18.80   &  1.57  &  2.71  &3.75 
             & 31.88 & 38.77 & 41.76 & 5.32 & 6.88 & 7.58
             \\ 
        $\text{OvSGTR}_\text{(Swin-T)}$ \cite{chen2024expanding}  &   (644k)      
           &  \textbf{9.94}  & \textbf{13.92}    &  \textbf{17.17}    &   \textbf{3.05} &   \textbf{4.03}  & \textbf{4.76} 
           & \textbf{37.12} & \textbf{44.10} & \textbf{47.09} &
           \textbf{8.49} & \textbf{10.22} & \textbf{11.07} \\
        $\text{OvSGTR}_\text{(Swin-B)}$ \cite{chen2024expanding}  &     
           &  \textbf{10.63}  & \textbf{14.93}    &  \textbf{18.36}    &   \textbf{3.01} &   \textbf{4.10}  & \textbf{4.99}
           & \textbf{38.72} & \textbf{45.71} & \textbf{48.51} 
           & \textbf{8.38} & \textbf{10.31} & \textbf{11.07} 
           \\   
   \bottomrule
    \end{tabular}
    }
\caption{Zero-shot performance of state-of-the-art methods on the VG150 test set. For the COCO Caption dataset, a language parser \cite{mao2018parser} has been used for extracting triplets from the caption. 
To prevent information leakage, we sampled 644k images from MegaSG, ensuring that the CLIP similarity of each sampled image with the VG test set remained below 0.9.
}
    \label{tab:mega_zeroshot}
\end{table*}

\noindent\textbf{Dataset Comparison.}
To verify the quality of MegaSG, we trained two state-of-the-art SGG models, \ie, $\text{VS}^3$~\cite{zhang2023learning} and OvSGTR~\cite{chen2024expanding}.
Table~\ref{tab:mega_zeroshot} reports the zero-shot performance of these two models trained on MegaSG.
From the result, MegaSG significantly improved the performance recall of OvSGTR from 22.79\% to 45.71\% (R@50, PredCls), offering a strong baseline to scale up SGG models.
Beyond the SGG task, the vast and diverse scenes offer a valuable resource for training and evaluating diffusion models based on scene graphs.

We compare the word cloud of VG and MegaSG in \cref{fig:wordcloud}. 
From the comparison in the word clouds, both the VG and MegaSG datasets contain similar high-frequency objects like ``person'', ``tree'', and ``man'', as well as common relationships such as ``on'' and ``near''.
However, the MegaSG dataset shows a wider variety of object types and relationship terms, suggesting it captures a wider range of visual semantics than the VG dataset.

%
%
%

\begin{figure*}[t]
    \centering

    \definecolor{ColorSDV1}{HTML}{0072B2} 
    \definecolor{ColorSDV2}{HTML}{E69F00} 
    \definecolor{ColorSDXL}{HTML}{009E73} 
    \definecolor{ColorOurs}{HTML}{D55E00} 

    \tikzset{
        SDVOneStyle/.style={mark=*, color=ColorSDV1, thick, dashed, mark options={solid, fill=ColorSDV1}, mark size=2pt},
        SDVTwoStyle/.style={mark=square*, color=ColorSDV2, thick, mark options={solid, fill=ColorSDV2}, mark size=2pt},
        SDXLStyle/.style={mark=triangle*, color=ColorSDXL, thick, dashed, mark options={solid, fill=ColorSDXL}, mark size=2pt},
        OursStyle/.style={mark=diamond*, color=ColorOurs, very thick, mark options={solid, fill=ColorOurs}, mark size=2pt},
    }

    \begin{subfigure}{0.48\textwidth}
        \centering
        \begin{tikzpicture}
            \begin{axis}[
                width=\textwidth,
                height=0.7\textwidth,
                xlabel={Scene Complexity ($\gamma=0$)},
                ylabel={FID},
                xmin=1, xmax=10,
                ymin=35, ymax=55,
                xtick={1,2,3,4,5,6,7,8,9,10},
                ytick={35,40,45,50,55},
                grid=both,
                label style={font=\footnotesize},
                tick label style={font=\footnotesize},
                legend style={font=\footnotesize, at={(0.5, 1.02)}, anchor=south, legend columns=2},
                legend cell align={left},
            ]
            \addplot[SDVOneStyle] coordinates {(1,43.29)(2,40.86)(3,39.34)(4,40.06)(5,41.07)(6,40.54)(7,42.86)(8,43.62)(9,44.93)(10,45.15)};
            \addlegendentry{SD v1.5}

            \addplot[SDVTwoStyle] coordinates {(1,48.32)(2,44.36)(3,40.87)(4,43.48)(5,46.18)(6,45.38)(7,49.74)(8,50.7)(9,52.62)(10,53.3)};
            \addlegendentry{SD v2.1}

            \addplot[SDXLStyle] coordinates {(1,49.34)(2,42.65)(3,38.92)(4,41.22)(5,43.68)(6,43.53)(7,47.6)(8,47.69)(9,49.62)(10,50.05)};
            \addlegendentry{SDXL}

            \addplot[OursStyle] coordinates {(1,40.27)(2,43.07)(3,41.94)(4,43.9)(5,46.84)(6,46.12)(7,50.43)(8,50.36)(9,51.88)(10,51.74)};
            \addlegendentry{Ours}

            \end{axis}
        \end{tikzpicture}
        \caption{FID Scores}
    \end{subfigure}
    \hfill
        \begin{subfigure}{0.48\textwidth}
        \centering
        \begin{tikzpicture}
            \begin{axis}[
                width=\textwidth,
                height=0.7\textwidth,
                xlabel={Scene Complexity ($\gamma=0$)},
                ylabel={ObjectRecall (\%)},
                xmin=1, xmax=10,
                ymin=50, ymax=100,
                xtick={1,2,3,4,5,6,7,8,9,10},
                ytick={50,60,70,80,90,100},
                grid=both,
                label style={font=\footnotesize},
                tick label style={font=\footnotesize},
                legend style={font=\footnotesize, at={(0.5, 1.02)}, anchor=south, legend columns=2},
                legend cell align={left},
            ]
            \addplot[SDVOneStyle] coordinates {(1,61.32)(2,78.29)(3,72.95)(4,70.39)(5,67.52)(6,63.91)(7,62.22)(8,59.23)(9,57.39)(10,56.06)};
            \addlegendentry{SD v1.5}

            \addplot[SDVTwoStyle] coordinates {(1,63.20)(2,80.09)(3,76.69)(4,73.66)(5,70.28)(6,68.71)(7,66.62)(8,64.55)(9,63.77)(10,62.25)};
            \addlegendentry{SD v2.1}

            \addplot[SDXLStyle] coordinates {(1,64.66)(2,85.31)(3,82.28)(4,80.28)(5,79.89)(6,77.12)(7,76.95)(8,74.79)(9,76.20)(10,75.28)};
            \addlegendentry{SDXL}

            \addplot[OursStyle] coordinates {(1,88.98)(2,95.06)(3,93.55)(4,92.18)(5,91.87)(6,90.92)(7,90.84)(8,89.53)(9,89.14)(10,89.25)};
            \addlegendentry{Ours}

            \end{axis}
        \end{tikzpicture}
        \caption{ObjectRecall}
    \end{subfigure}
    \vfill 
    \begin{subfigure}{0.48\textwidth}
        \centering
        \begin{tikzpicture}
            \begin{axis}[
                width=\textwidth,
                height=0.7\textwidth,
                xlabel={Scene Complexity ($\gamma=0$)},
                ylabel={RelationRecall (\%)},
                xmin=1, xmax=10,
                ymin=20, ymax=90,
                xtick={1,2,3,4,5,6,7,8,9,10},
                ytick={20,30,40,50,60,70,80,90},
                grid=both,
                label style={font=\footnotesize},
                tick label style={font=\footnotesize},
                legend style={font=\footnotesize, at={(0.5, 1.02)}, anchor=south, legend columns=2},
                legend cell align={left},
            ]
            \addplot[SDVOneStyle] coordinates {(1,70.64)(2,57.26)(3,49.00)(4,45.93)(5,42.64)(6,38.41)(7,36.73)(8,33.49)(9,32.10)(10,30.39)};
            \addlegendentry{SD v1.5}

            \addplot[SDVTwoStyle] coordinates {(1,76.62)(2,60.64)(3,52.46)(4,48.07)(5,44.08)(6,40.82)(7,39.12)(8,36.29)(9,35.28)(10,33.50)};
            \addlegendentry{SD v2.1}

            \addplot[SDXLStyle] coordinates {(1,80.76)(2,67.28)(3,59.32)(4,53.52)(5,50.87)(6,48.61)(7,46.55)(8,42.80)(9,42.44)(10,42.85)};
            \addlegendentry{SDXL}

            \addplot[OursStyle] coordinates {(1,73.16)(2,72.95)(3,67.47)(4,65.32)(5,62.51)(6,60.59)(7,58.84)(8,58.07)(9,56.78)(10,56.24)};
            \addlegendentry{Ours}

            \end{axis}
        \end{tikzpicture}
        \caption{RelationRecall}
    \end{subfigure}
    \hfill 
    \begin{subfigure}{0.48\textwidth}
        \centering
        \begin{tikzpicture}
            \begin{axis}[
                width=\textwidth,
                height=0.7\textwidth,
                xlabel={Scene Complexity ($\gamma=0$)},
                ylabel={SGScore (\%)},
                xmin=1, xmax=10,
                ymin=40, ymax=90,
                xtick={1,2,3,4,5,6,7,8,9,10},
                ytick={40,50,60,70,80,90},
                grid=both,
                label style={font=\footnotesize},
                tick label style={font=\footnotesize},
                legend style={font=\footnotesize, at={(0.5,1.02)}, anchor=south, legend columns=2},
                legend cell align={left},
            ]
            \addplot[SDVOneStyle] coordinates {(1,65.98)(2,67.77)(3,60.98)(4,58.16)(5,55.08)(6,51.16)(7,49.47)(8,46.36)(9,44.75)(10,43.23)};
            \addlegendentry{SD v1.5}

            \addplot[SDVTwoStyle] coordinates {(1,69.91)(2,70.36)(3,64.57)(4,60.86)(5,57.18)(6,54.77)(7,52.87)(8,50.42)(9,49.52)(10,47.88)};
            \addlegendentry{SD v2.1}

            \addplot[SDXLStyle] coordinates {(1,72.71)(2,76.29)(3,70.8)(4,66.9)(5,65.38)(6,62.86)(7,61.75)(8,58.79)(9,59.32)(10,59.06)};
            \addlegendentry{SDXL}

            \addplot[OursStyle] coordinates {(1,81.07)(2,84.01)(3,80.51)(4,78.75)(5,77.19)(6,75.75)(7,74.84)(8,73.80)(9,72.96)(10,72.74)};
            \addlegendentry{Ours}

            \end{axis}
        \end{tikzpicture}
        \caption{SGScore}
    \end{subfigure}    
    \caption{Comparison of FID, ObjectRecall,  RelationRecall, and SGScore for models SD v1.5, SD v2.1, SDXL, and Ours across different scene complexity levels.
    (a) FID scores show relatively stable image quality, while (b) ObjectRecall and (c) RelationRecall indicate a consistent decline in factual consistency with increasing scene complexity. (d) SGScore demonstrates the overall advantage of our approach in maintaining higher factual consistency, particularly in complex scenes.
    }
    \label{fig:complexity_impact}
\end{figure*}
\section{Experiments}
\label{app:exp}
\subsection{Experimental Setup} 
\label{sec:supp_exp}
\textbf{Models.} 
We evaluate several popular open-source diffusion models, including 
Composable~\cite{liu2022compositional}, 
Structured~\cite{feng2023training}, 
SD v1.5~\cite{rombach2022high} 
(checkpoint: \emph{runwayml/stable-diffusion-v1-5}), 
SD v2.1~\cite{rombach2022high}
(checkpoint: \emph{stabilityai/stable-diffusion-2-1}), 
PixArt-$\alpha$~\cite{chen2024pixart} (checkpoint: \emph{PixArt-alpha/PixArt-XL-2-1024-MS}), 
SD3~\cite{esser2024scaling} (checkpoint: \emph{stabilityai/stable-diffusion-3-medium-diffusers}), 
SD3.5~\cite{esser2024scaling} (checkpoint: \emph{stabilityai/stable-diffusion-3.5-large}), 
SDXL~\cite{podellelbdmpr24} (checkpoint: \emph{stabilityai/stable-diffusion-xl-base-1.0}), 
and LLM-based methods such as RPG~\cite{yang2024mastering}. 
We use \texttt{diffusers}~\cite{von-platen-etal-2022-diffusers} or official code to benchmark these models.

\noindent\textbf{Datasets.}
We benchmark models on the widely used Visual Genome (VG) and the proposed MegaSG dataset.
\begin{itemize}
    \item VG  consists of 108k images annotated by human. Following SG2Im~\cite{johnson2018image}, it has been split into training set ( 62,565), validation set (5,506), and test set (5,088\footnote{we use official code to obtain 5,096 images for test.}) images for scene graph-based image generation. 
    \item MegaSG comprises 1 million images annotated using Gemini 1.5 Flash. Relationships with a frequency below 100 are filtered out, and synonyms are merged by a large language model (LLM).  
\end{itemize}

\noindent\textbf{Metrics.}
We employ common metrics and the proposed SGScore.
\begin{itemize}
    \item Inception Score (IS)~\cite{salimans2016improved}: Measures the realism of generated images using a pre-trained Inception-V3~\cite{szegedy2016rethinking} network.
    \item Fr\'{e}chet Inception Distance (FID)~\cite{heusel2017gans}: Assesses the similarity between generated and real images by measuring the distance between the distributions of their feature representations.
    \item CLIPScore~\cite{hessel2021clipscore}: Evaluates the semantic alignment between generated images and corresponding text using the CLIP model~\cite{radford2021learning}.
    \item SGScore measure the factual consistency in terms of object recall and relation recall. We use $\alpha=0.5$ in Eq. (4) of Section 3.2 to give a balanced measurement.  
\end{itemize}

\begin{figure*}[t]
    \centering
    \includegraphics[width=\textwidth]{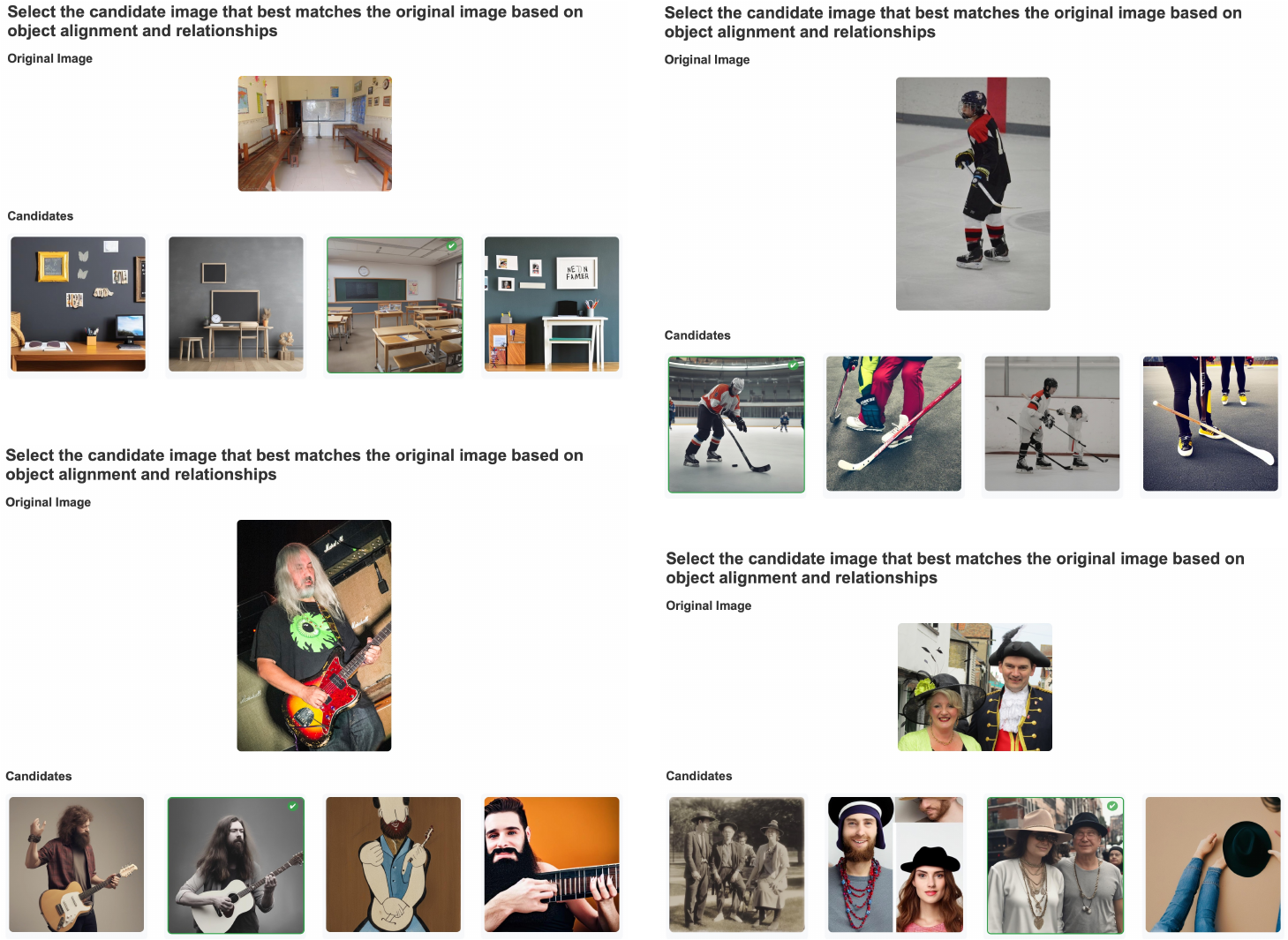}
    \caption{Example questions presented to human annotators.}
    \label{fig:human-question}
\end{figure*}
\noindent\textbf{Scene Graph Representation.} 
For text-to-image (T2I) models that condition on a sentence, we encode scene graphs in the format 
\texttt{\{subject\} \{predicate\} \{object\}} (e.g., \texttt{cat sitting on desk, dog near chair}). 
The prompt used to convert a scene graph into a consistent description (\ie, the scene composition described in Section~\ref{sec:feedback}) is provided in Table~\ref{tab:prompt_sg_comp}.
\begin{table*}[t]
\centering
\definecolor{colorbg@intro}{RGB}{240,240,240} 
\begin{tikzpicture}[
    node distance=1cm and 0.5cm,
    auto,
    description/.style={
        rectangle, rounded corners=2pt, draw, thick, fill=colorbg@intro,
        text width=\textwidth,
        align=left,
        scale=1.0
    },
]
\node[description] {%
\textcolor{blue}{messages} = [
\{ \texttt{"role"}: \texttt{"user"}, \texttt{"content"}: ``You are an AI assistant tasked with converting scene graphs into descriptive text prompts for image generation using diffusion models. Given the input scene graph, generate a detailed text description adhering to the following instructions:\\ 
\quad **Instructions:**\\
\quad 1. **Scene Setup:** Begin by describing the overall setting or background of the scene. If no explicit background object exists, infer a suitable one from the present objects and relationships.\\ 
\quad 2. **Object Placement:** Introduce each object from the `objects' list. When describing their placement, utilize the `relationships' to accurately depict their positions relative to each other and the scene.\\ 
\quad 3. **Relationship Emphasis:** Vividly express the relationships between objects using descriptive language. Avoid simply stating the relationship; instead, showcase it through the objects' placement, appearance, or actions.\\ 
\quad 4. **Image-ability:** Craft the description to be easily translatable into a visual representation. Use evocative language that captures the essence of the scene and guides the diffusion model.\\ 
\quad 5. **Conciseness and Clarity:** Be concise and avoid unnecessary details. Ensure the language is unambiguous and accurately reflects the scene graph information.\\ 
\quad **Output Format:** Provide the output as a JSON object with a single key \texttt{"description"} and the value being the generated text description.\\ 
\quad **Input Scene Graph (JSON format):**\\ 
\quad \texttt{```json}  \{scene\_graph\}
\texttt{```}\\
\quad **Output:**"
\}
]; \\ [2ex]

\textcolor{blue}{example\_input} = [ `objects': [`cymbal.1', `cymbal.2', `drum.3', `guitar.4', `microphone.5', `person.6'], `relationships': [\{`source': `person.6', `target': `guitar.4', `relation': `playing'\}, \{`source': `person.6', `target': `microphone.5', `relation': `using'\}, \{`source': `guitar.4', `target': `cymbal.1', `relation': `near'\}, \{`source': `guitar.4', `target': `cymbal.2', `relation': `near'\}, \{`source': `guitar.4', `target': `drum.3', `relation': `near'\}]]; \\ [1ex]

\textcolor{blue}{example\_output}= ``A musician stands on a stage, the bright lights reflecting off their instruments. They are playing a guitar, their fingers dancing across the strings. A microphone is positioned in front of them, capturing their performance. The guitar rests near a pair of cymbals, and a drum sits nearby, adding to the musical ensemble.''
};
\end{tikzpicture}
\caption{Prompt for scene composition: translating a concise scene graph into a coherent and descriptive text.}
\label{tab:prompt_sg_comp}
\end{table*}

\noindent\textbf{Scene Complexity.}
We define the scene complexity of a scene graph $G = (V, E)$ as:
\begin{equation}
    C(G) = \gamma \cdot |V| + (1 - \gamma) \cdot |E|, 
    \label{eq:comp_suppl}
\end{equation}
where $\gamma$ is a weighting factor. The three levels of complexity are defined as follows:
\begin{itemize}
    \item \textbf{Simple}: $1 \leq C(G) \leq 3$, typically involving 2–3 objects and no more than 3 relationships in the scene.
    \item \textbf{Medium}: $4 \leq C(G) \leq 7$, characterized by a denser arrangement of objects and relationships.
    \item \textbf{Hard}: $C(G) \geq 8$, representing the most challenging cases with highly dense objects and intricate relationships.
\end{itemize}

\noindent\textbf{LLM.} In addition to utilizing Gemini 1.5 Flash, we also present results using GPT-4o~\cite{achiam2023gpt}, Qwen-VL-Max~\cite{wang2024qwen2}, and LLaVA 1.5~\cite{liu2024llavanext} to evaluate the robustness of the proposed \emph{SGScore}.

\noindent\textbf{IP-Adapter.} We use the official implementation in \texttt{diffusers}~\cite{von-platen-etal-2022-diffusers}, with $\lambda_0$ and $\lambda_1$ (in Eq. (5) of Sec. 4) empirically set to 0.5.

\subsection{Evaluation of Scene-Bench}
\label{sec:supp_eval_sgbench}
\textbf{Impact of Scene Complexity.} 
To examine how scene complexity affects model performance, 
we analyzed FID and SGScore for SD v1.5, SD v2.1, SDXL, and our model across various complexity levels (see \cref{fig:complexity_impact}).

As scene complexity increases (\ie, with more objects and relationships), 
we observe a consistent decline in SGScore across all models. 
This suggests that, with greater complexity, 
the models struggle to accurately represent the expected scene graphs. 
The decreasing SGScore highlights the challenge of maintaining factual consistency in complex scenes.
However, our model demonstrates a notable improvement over the other models by consistently achieving a higher SGScore across all complexity levels, 
particularly through maintaining stable and high object recall. 
This suggests that our model is more effective at preserving factual consistency even in complex scenes.

Interestingly, FID scores remain stable across complexity levels, 
indicating that image quality does not degrade significantly with complexity. 
This stability implies that while models retain visual fidelity, 
they encounter difficulties modeling intricate object relationships and interactions in complex scenes. Therefore, even as images appear visually coherent, the factual accuracy, as measured by SGScore, declines with increased scene complexity.

\begin{table}[t]
    \centering
    \resizebox{\columnwidth}{!}{
    \begin{tabular}{ccccc}
    \toprule 
        IP-Adapter & Ref. Img. & ObjectRecall & RelationRecall & SGScore \\
    \midrule 
        \xmark &   \xmark  & 75.45 & 48.84 &  62.14 \\
        \cmark & \xmark & 70.91 &  49.83  & 60.37 \\
        \cmark & \cmark & $\mathbf{79.93}$ & $\mathbf{53.97}$ & $\mathbf{66.95}$ \\
    \bottomrule 
    \end{tabular}
    }
    \caption{Comparison of ObjectRecall, RelationRecall, and SGScore with and without the reference image in the IP-Adapter setup. ``Ref. Img.'' denotes the reference image.
    }
    \label{tab:ablation_ip}
\end{table}
\subsection{Evaluation of Scene Graph Feedback}
\textbf{Additional Ablation Study.} 
\label{sec:ablation_ip}
To evaluate the effectiveness of the scene graph feedback, particularly considering the additional parameters introduced by the IP-Adapter,
we conducted another ablation study. 
In this experiment, we set $\lambda_1 = 0$ in \cref{eq:feedback} of \cref{sec:feedback}, 
meaning the IP-Adapter processes only the initial generated image, without incorporating the reference image derived from the missing graph.

As shown in \cref{tab:ablation_ip}, introducing the IP-Adapter alone (row 2 vs. row 1) does not improve the factual consistency of generated images. 
However, incorporating the reference image (row 3) significantly enhances ObjectRecall, RelationRecall, and SGScore, demonstrating the importance of scene graph feedback.

\subsection{Human Evaluation}
\label{sec:supp_human}
\begin{figure}[t]
    \centering
    \includegraphics[width=0.985\columnwidth]{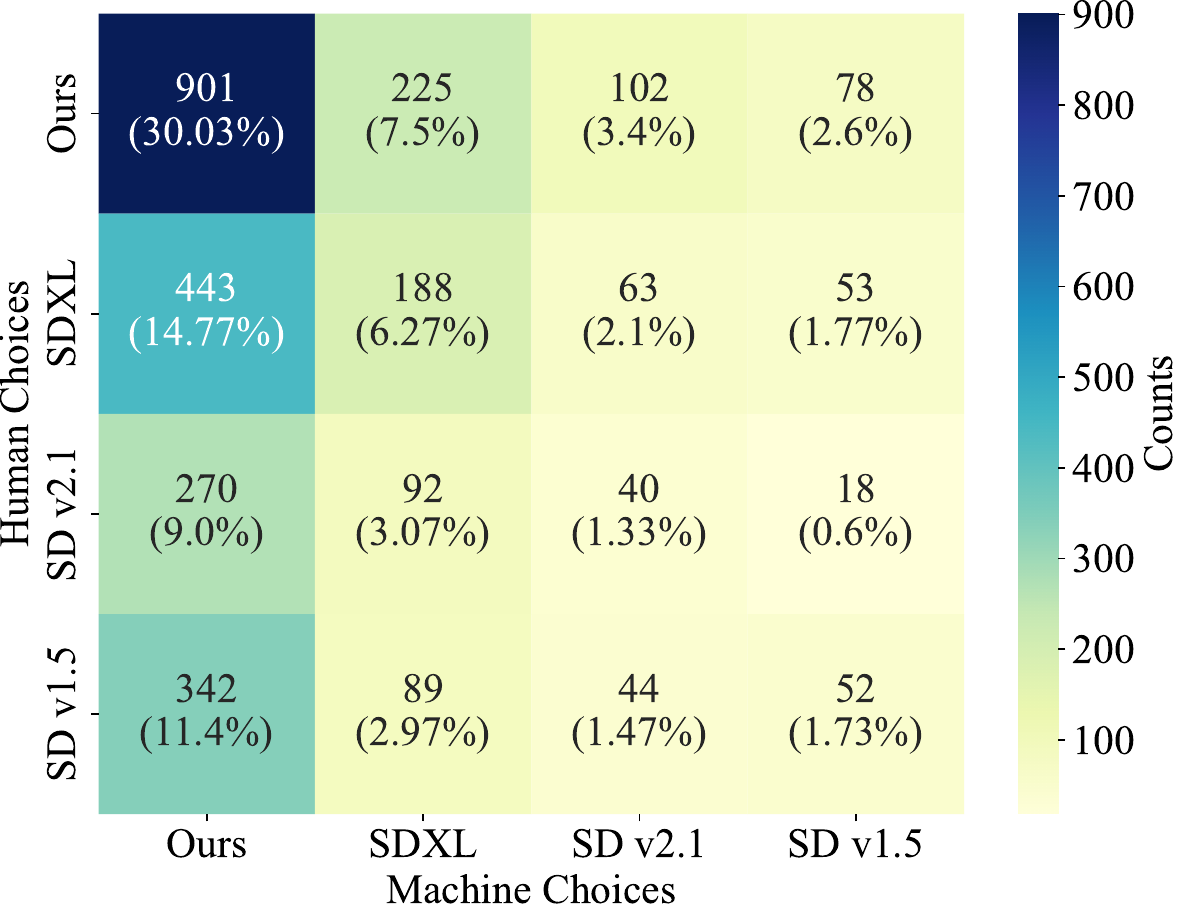}
    \caption{Confusion matrix showing the comparison of human choices against machine choices based on SGScore.}
    \label{fig:confusion}
\end{figure}
We conducted a human evaluation to assess the effectiveness of SGScore in verifying factual consistency.
Specifically, three human annotators were instructed to select the candidate image that best aligns with the original image regarding object presence and relationship accuracy. 
We randomly sampled 1,000 images and selected corresponding generated images from four models: SD v1.5, SD v2.1, SDXL, and Ours (SDXL). 
Model identities were hidden from the annotators to avoid bias.
\cref{fig:human-question} illustrates the annotation interface.

\section{Additional Results}
\label{app:add}
\subsection{Multimodal LLMs for SGScore}
\label{sec:supp_llm}
\begin{figure}[t]
    \centering
    \includegraphics[width=\columnwidth]{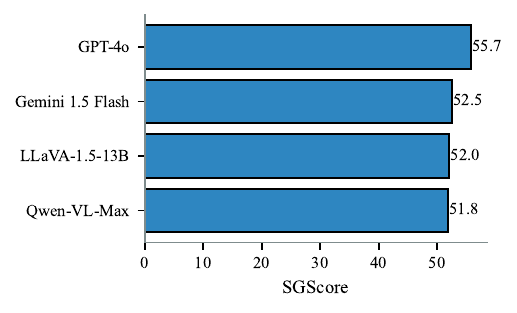}
    \caption{Performance comparison of M-LLMs on VG test set (images are generated by SD v1.5).
    }
    \label{fig:sgscore_comparison_mllm}
\end{figure}

We evaluate the performance of different multimodal LLMs on SGScore, 
including Gemini 1.5 Flash~\cite{reid2024gemini}, GPT-4o~\cite{achiam2023gpt}, Qwen-VL-Max~\cite{wang2024qwen2}, and LLaVA 1.5~\cite{liu2024llavanext}, as shown in \cref{fig:sgscore_comparison_mllm}. 
The results show minimal discrepancies across models, indicating that SGScore is insensitive to the specific choice of state-of-the-art multimodal LLMs for the same evaluation. 
However, the visual reasoning capability of these models remains important.
Considering cost-effectiveness, we recommend Gemini 1.5 Flash, which offers excellent multimodal reasoning performance at a significantly lower price~\cite{google2024gemini}, as indicated in \cref{sec:comp_cost}.

\subsection{More Qualitative Results}
\begin{figure*}[t]
    \centering
    \definecolor{personcolor}{HTML}{FF6666}  
    \definecolor{objectcolor}{HTML}{66CCFF}  
    \newcommand{\imggap}{-12.5mm}
    \resizebox{\textwidth}{!}
    {
    \begin{tikzpicture}
        \node (sg00) [align=center] at(0, 0) {
            \includegraphics[width=2.5cm]{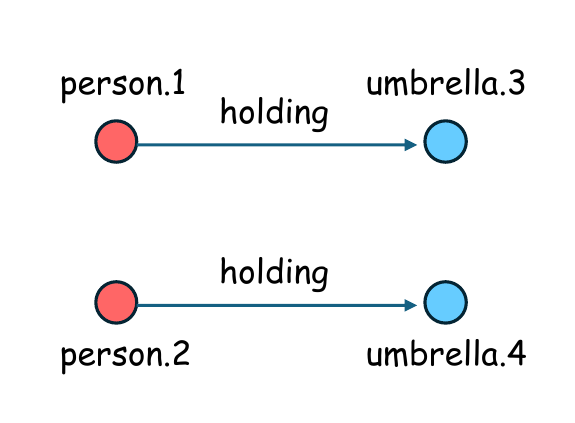}
        };
        \node (sg02) [align=center] at(0, -2.4) {
            \includegraphics[width=2.5cm]{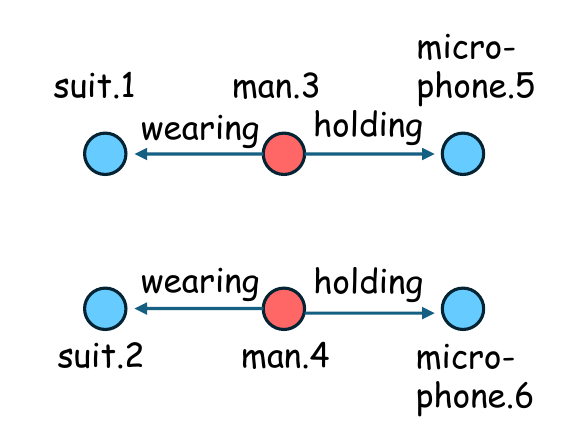}
        };  
        \node (sg04) [align=center] at(0, -4.8) {
            \includegraphics[width=2.5cm]{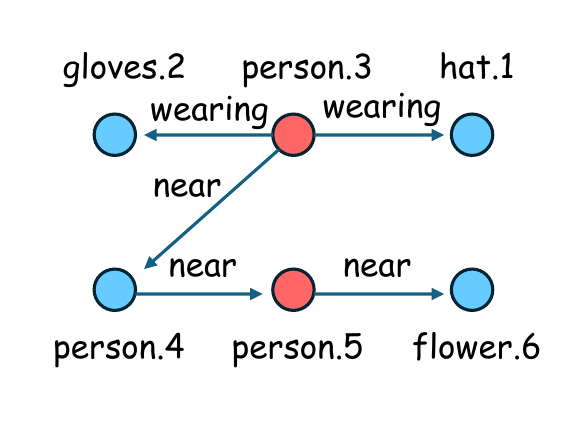}
        };     
        
        \node (n01) [align=center, right=of sg00, xshift=-12mm, yshift=0mm]  {
            \includegraphics[width=2cm, height=2cm]{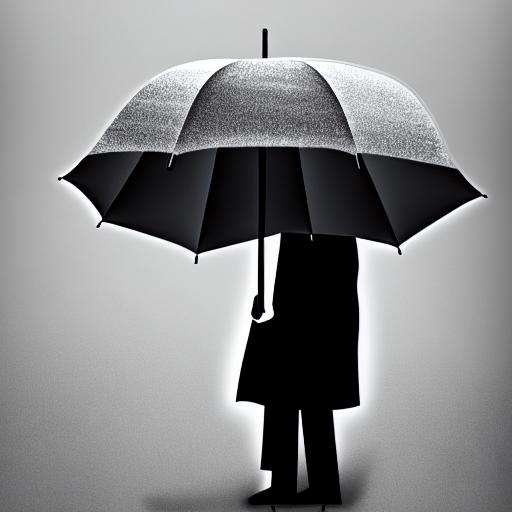}
        };
        \node [below=of n01, yshift=12mm]  {\small 50.0};
        
        \node (n02) [align=center, right=of n01, xshift=\imggap]     {
            \includegraphics[width=2cm, height=2cm]{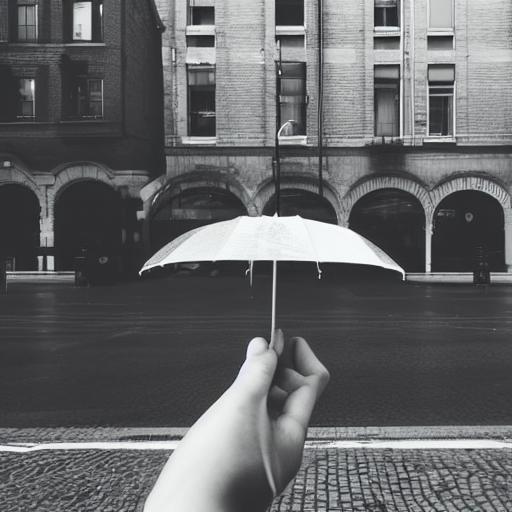}
        };
        \node [below=of n02, yshift=12mm]  {\small 50.0};
        
        \node (n03) [align=center, right=of n02, xshift=\imggap]  {
            \includegraphics[width=2cm, height=2cm]{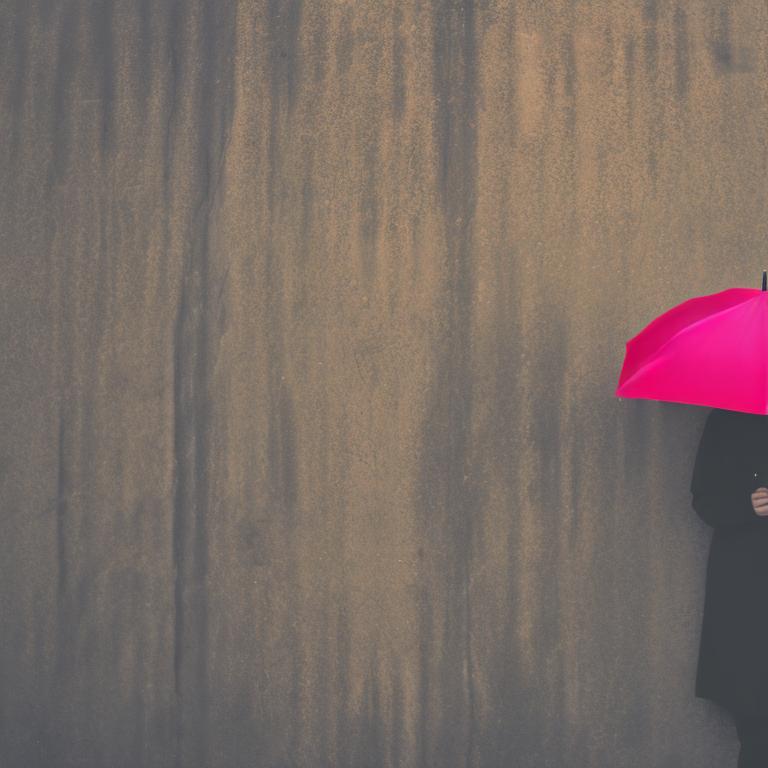} 
        };
        \node [below=of n03, yshift=12mm]  {\small 50.0};
        
        \node (n04) [align=center, right=of n03, xshift=\imggap]    {
            \includegraphics[width=2cm, height=2cm]{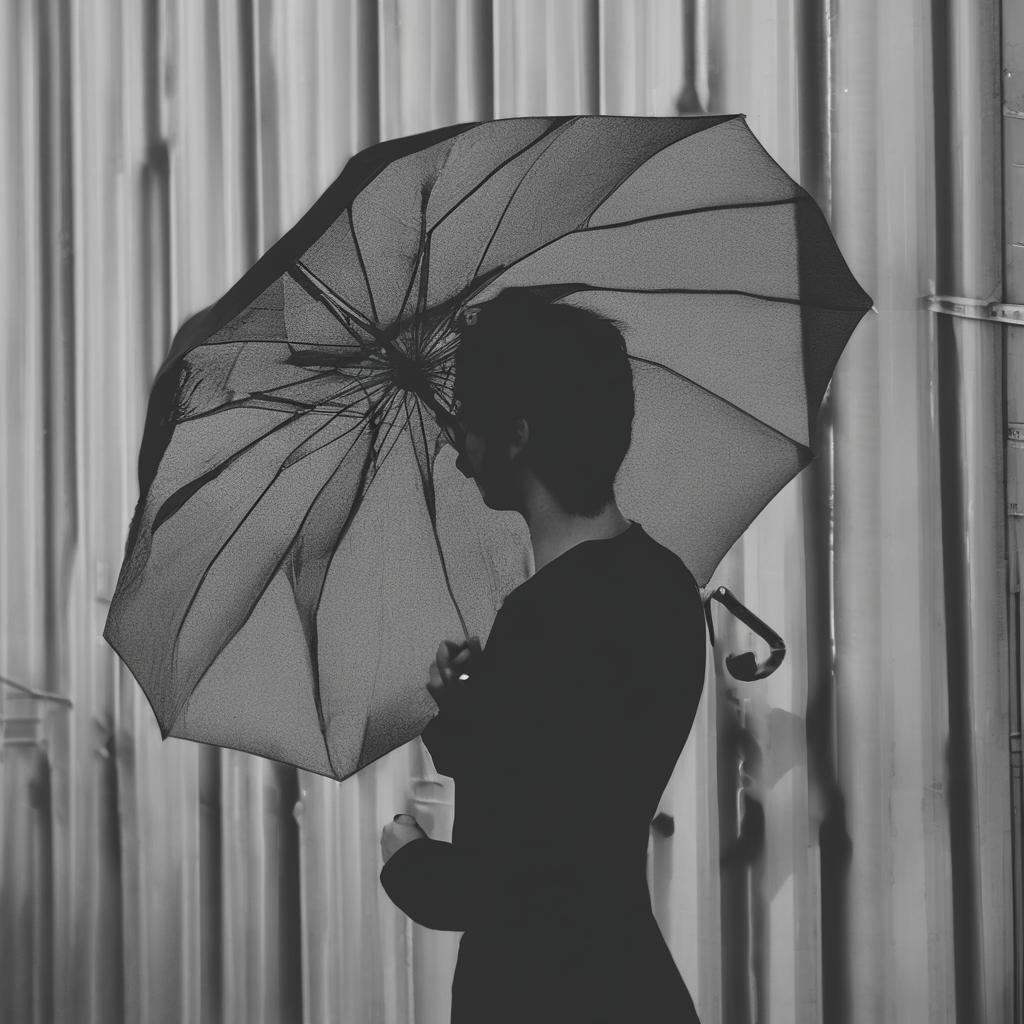} 
        };
        \node [below=of n04, yshift=12mm]  {\small 50.0};
        
        \node (n05) [align=center, right=of n04, xshift=\imggap]    {
            \includegraphics[width=2cm, height=2cm]{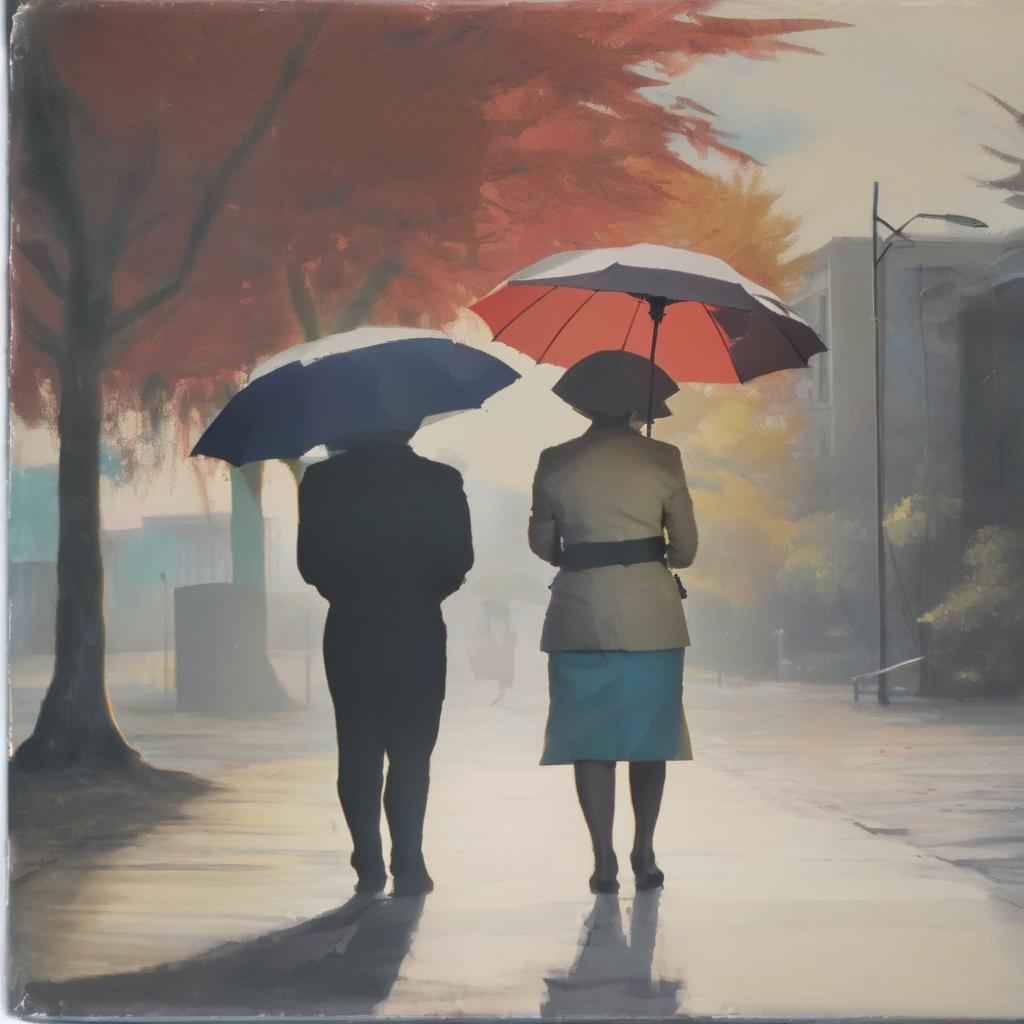} 
        };
        \node [below=of n05, yshift=12mm]  {\small 100.0};
        
        \node (t01) [above=of n01.north, yshift=-10.8mm] {Composable};
        \node [above=of n02.north, yshift=-10mm] {SD v1.5};
        \node [above=of n03.north, yshift=-10mm] {SD v2.1};
        \node [above=of n04.north, yshift=-10mm] {SDXL};
        \node [above=of n05.north, yshift=-10mm] {Ours};
        \node [left=of t01, xshift=5mm] {Scene Graph};
        
        \node (n21) [align=center, right=of sg02, xshift=-12mm, ]  {
            \includegraphics[width=2cm, height=2cm]{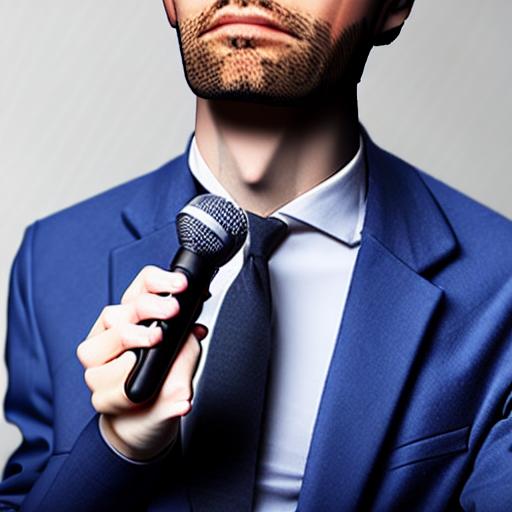}
        };
        \node [below=of n21, yshift=12mm]  {\small 50.0};
        
        \node (n22) [align=center, right=of n21, xshift=\imggap]     {
            \includegraphics[width=2cm, height=2cm]{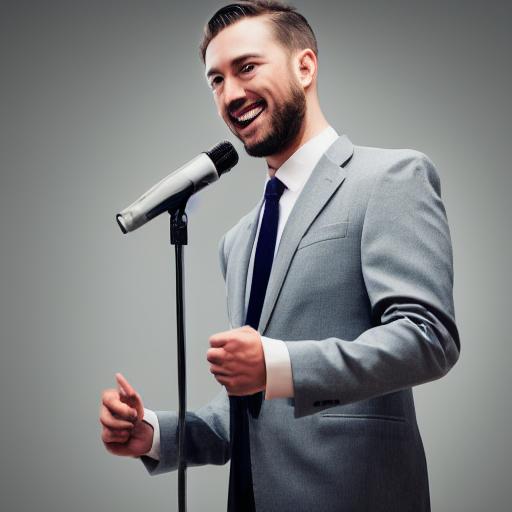}
        };
        \node [below=of n22, yshift=12mm]  {\small 50.0};
        
        \node (n23) [align=center, right=of n22, xshift=\imggap]  {
            \includegraphics[width=2cm, height=2cm]{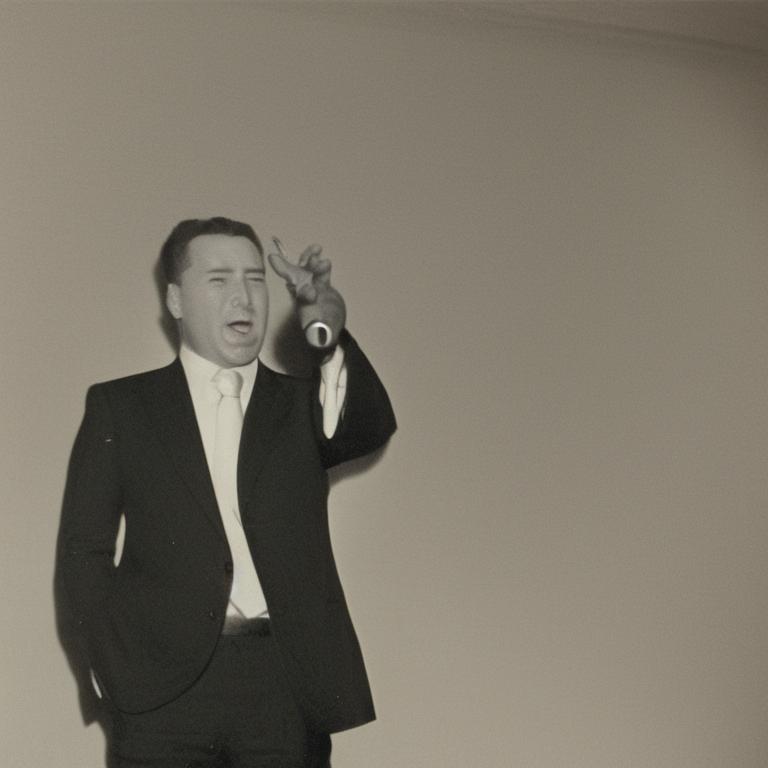}
        };
        \node [below=of n23, yshift=12mm]  {\small 50.0};
        
        \node (n24) [align=center, right=of n23, xshift=\imggap]    {
            \includegraphics[width=2cm, height=2cm]{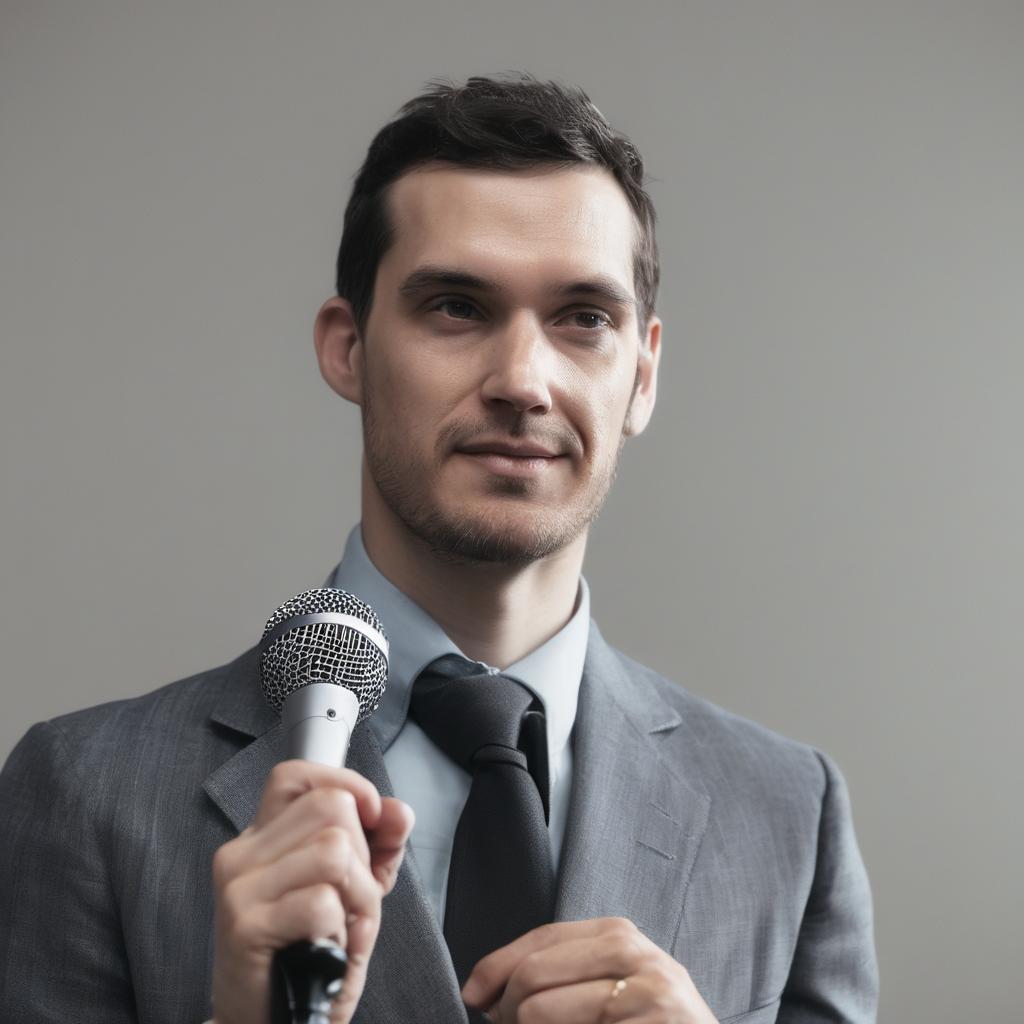}
        };
        \node [below=of n24, yshift=12mm]  {\small 50.0};
        
        \node (n25) [align=center, right=of n24, xshift=\imggap]    {
            \includegraphics[width=2cm, height=2cm]{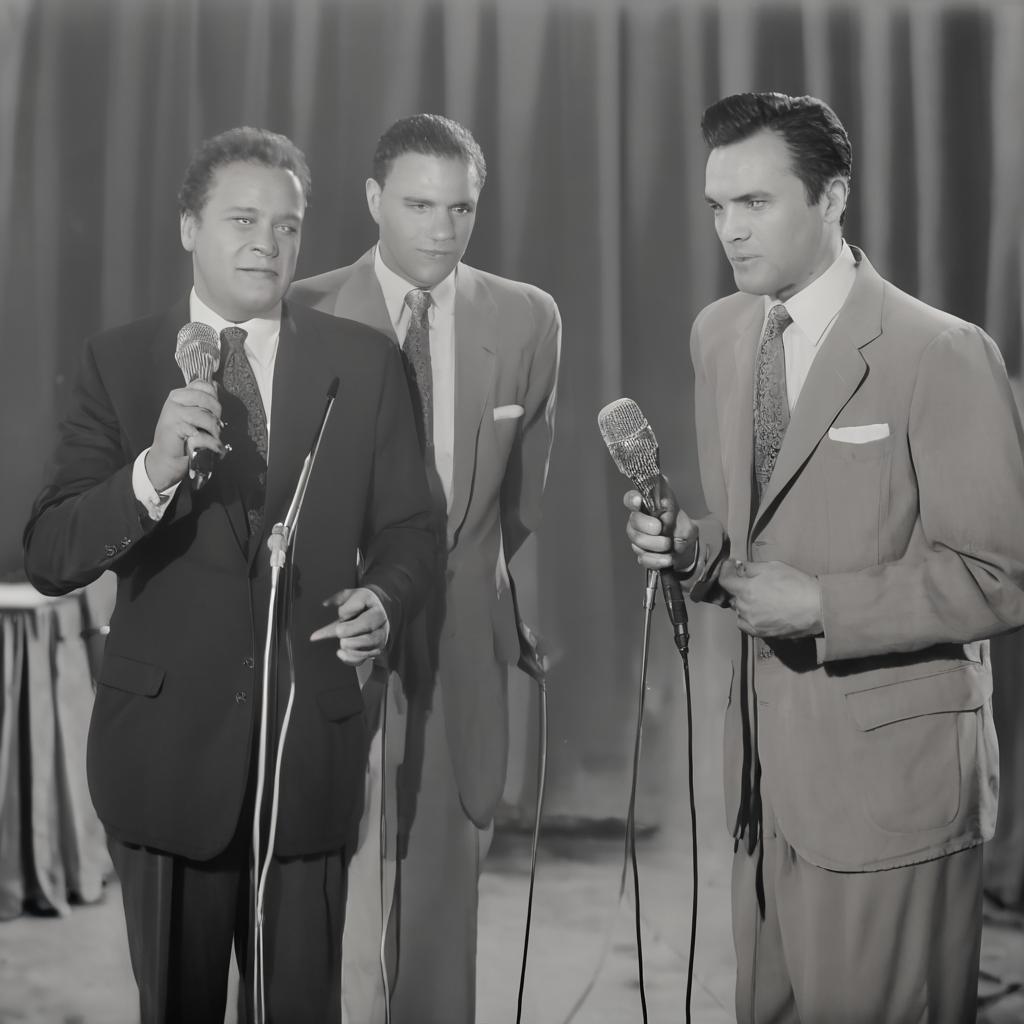}
        };
        \node [below=of n25, yshift=12mm]  {\small 100.0};

        \node (n41) [align=center, right=of sg04, xshift=-12mm, ]  {
            \includegraphics[width=2cm, height=2cm]{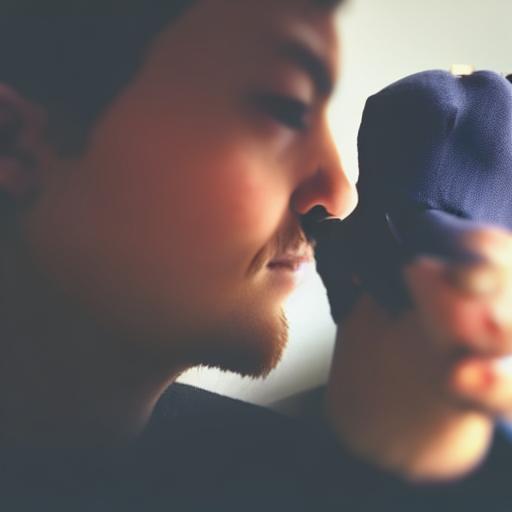}
        };
        \node [below=of n41, yshift=12mm]  {\small 25.0};
        
        \node (n42) [align=center, right=of n41, xshift=\imggap]     {
            \includegraphics[width=2cm, height=2cm]{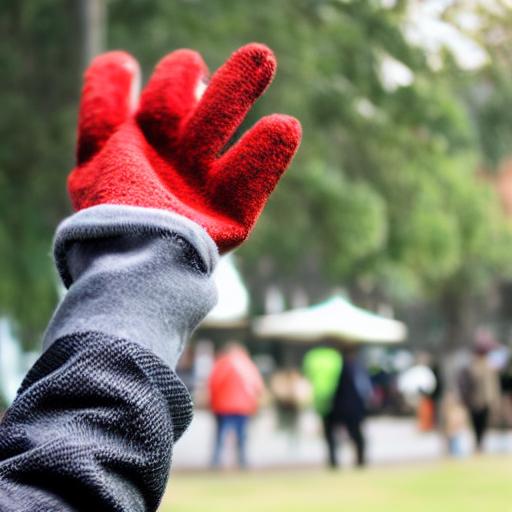}
        };
        \node [below=of n42, yshift=12mm]  {\small 16.6};
        
        \node (n43) [align=center, right=of n42, xshift=\imggap]  {
            \includegraphics[width=2cm, height=2cm]{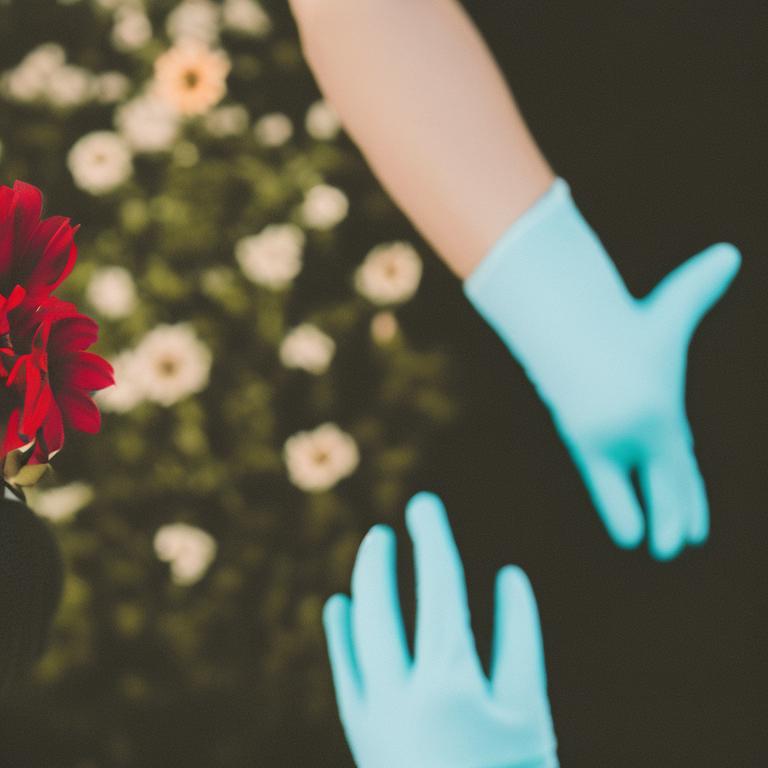}
        };
        \node [below=of n43, yshift=12mm]  {\small 16.6};
        
        \node (n44) [align=center, right=of n43, xshift=\imggap]    {
            \includegraphics[width=2cm, height=2cm]{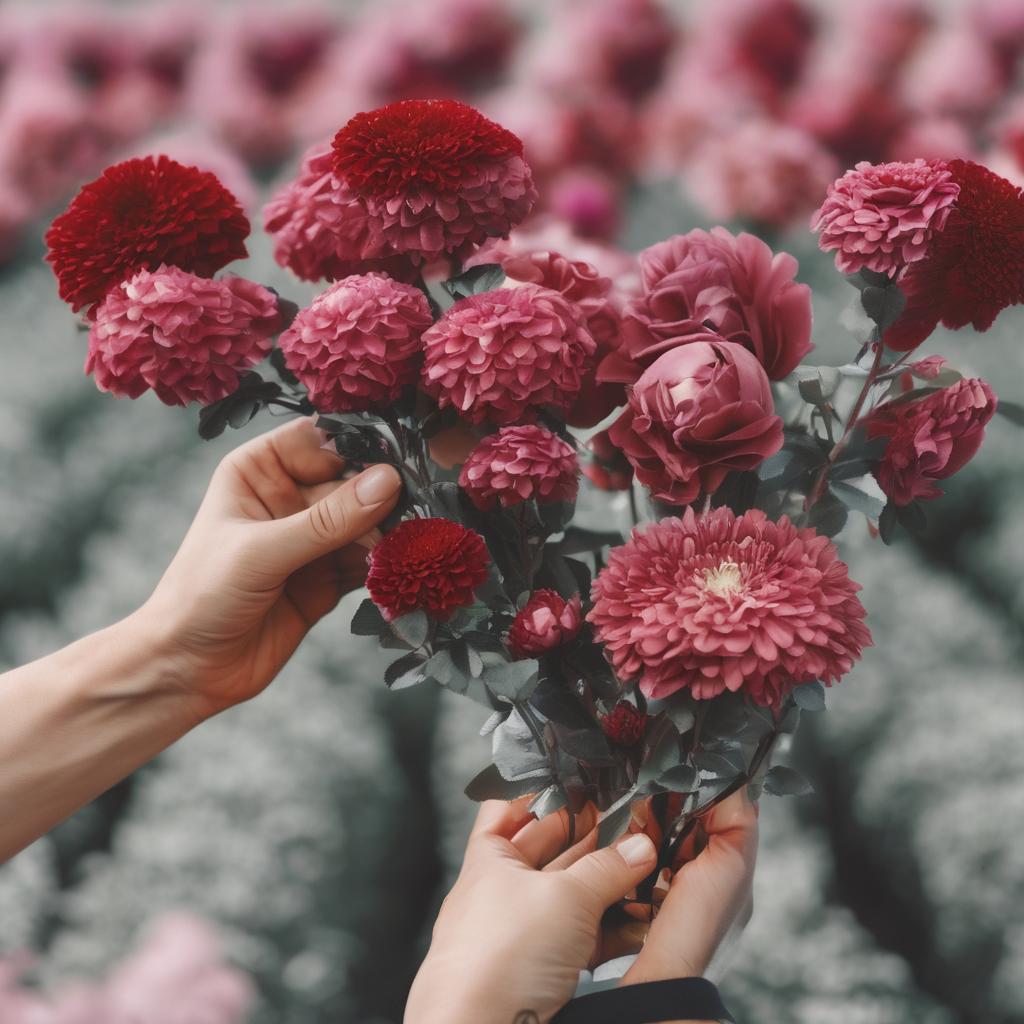}
        };
        \node [below=of n44, yshift=12mm]  {\small 41.6};
        
        \node (n45) [align=center, right=of n44, xshift=\imggap]    {
            \includegraphics[width=2cm, height=2cm]{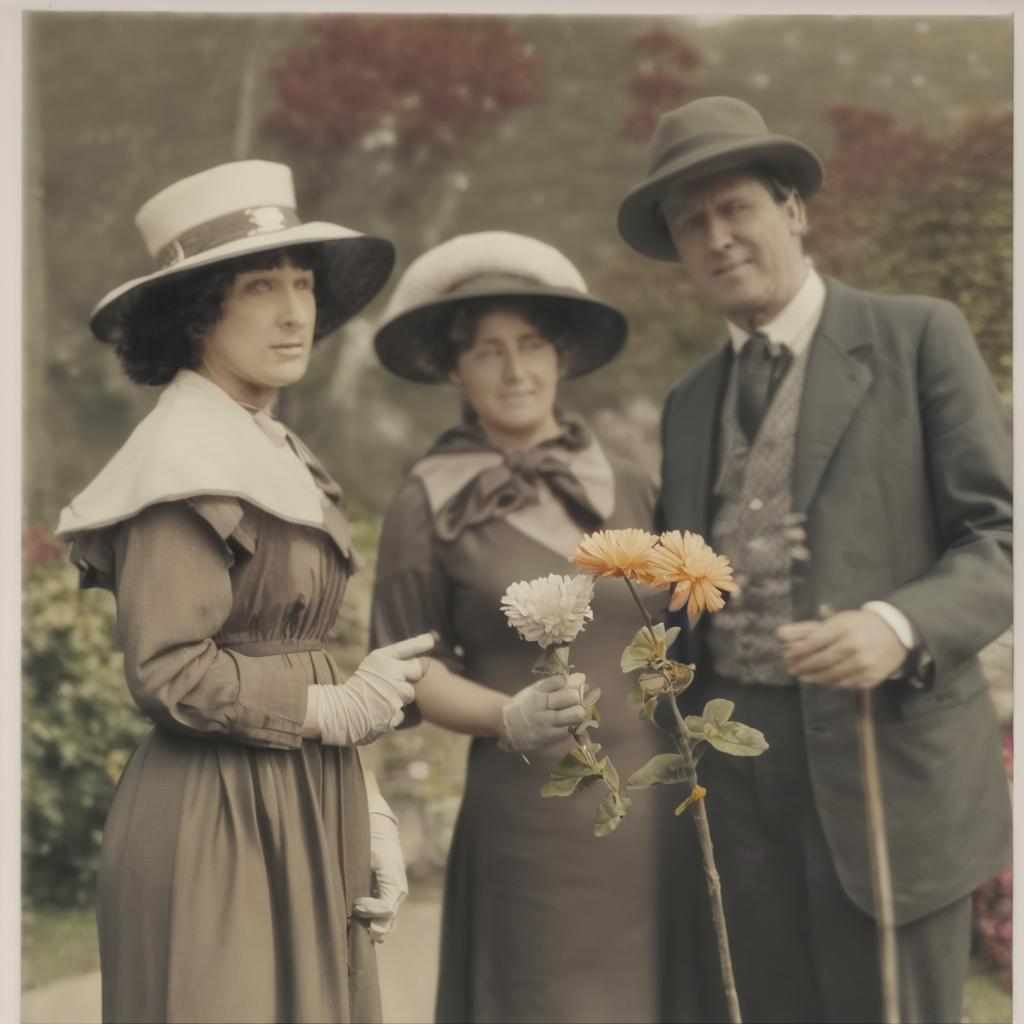}
        };
        \node [below=of n45, yshift=12mm]  {\small 100.0};    
    \end{tikzpicture}
    }
\caption{Comparison of Scene Graph-based Image Generation across Different Models. 
Each row displays a unique scene graph used as input for image generation. We present the \textbf{SGScore} below each generated image to quantify the consistency between the scene graph and the generated output. 
}
    \label{fig:extra_qualitative_results}
\end{figure*}
We present additional qualitative examples in \cref{fig:extra_qualitative_results}, 
highlighting our model's ability to handle complex scenes with multiple instances of the same object or relationship categories (1st row) 
as well as intricate indoor or outdoor scenes (2nd and 3rd rows).
These results demonstrate our framework's effectiveness in generating scene graph–guided images with high semantic fidelity.
%
%
\section{Discussion}
\subsection{Sensitivity Analysis of SGScore}

\begin{figure}[t]
    \centering
    \includegraphics[width=0.985\columnwidth]{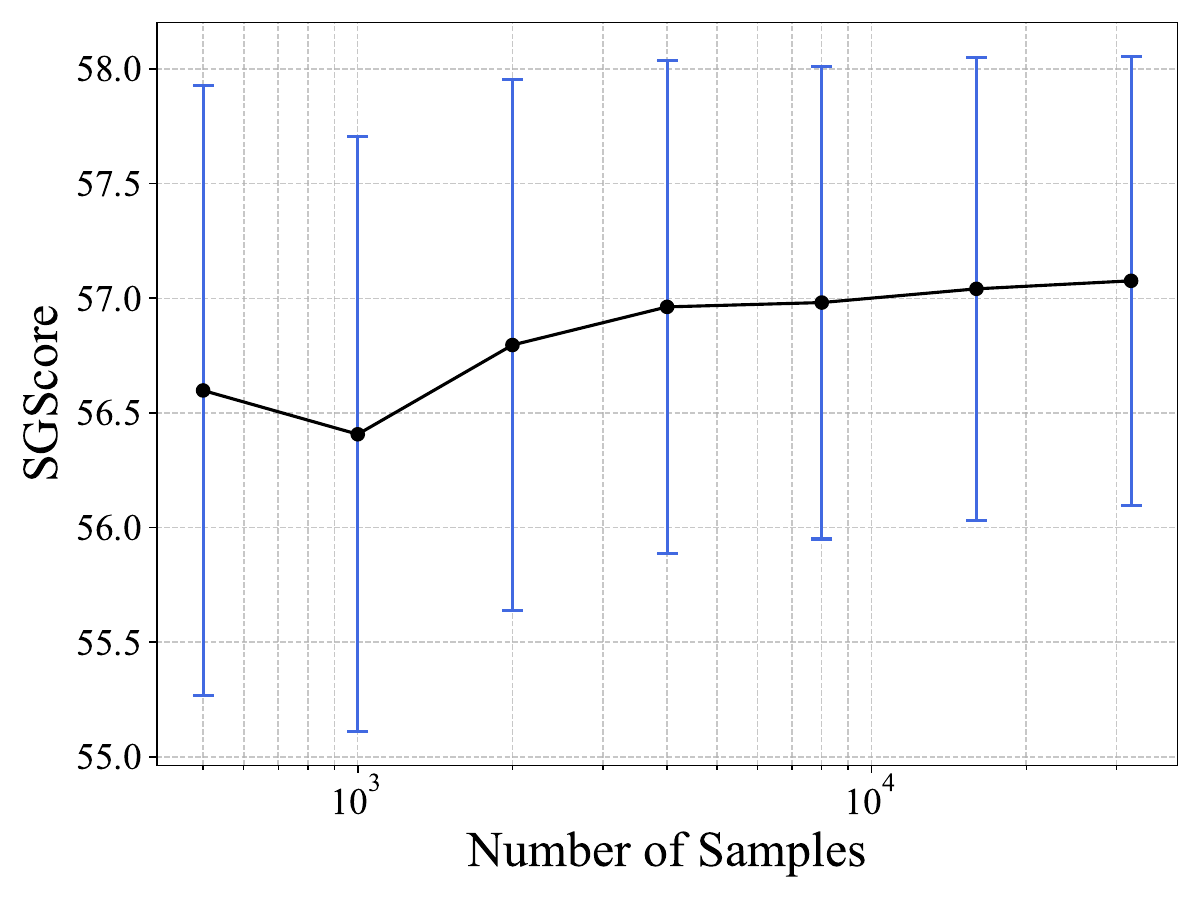}
    \caption{Mean and standard deviation (std.) of SGScore across varying numbers of samples. 
    For each sample size, the image generation process is repeated with different random seeds using SD v1.5 to compute the mean and std. of SGScore.
    }
    \label{fig:errorbar1}
\end{figure}

To test the scalability and sensitivity of SGScore, we randomly sample subsets (\eg, with size 500, 1k, 2k, 4k, $\cdots$, 32k, \etc) from the MegaSG to compute the SGScore on images generated by SD v1.5. 
As shown in \cref{fig:errorbar1}, the mean SGScore remains relatively stable across sample sizes, with only slight variations observed.
Additionally, the standard deviation decreases as the sample size increases, demonstrating that the metric becomes more reliable and less sensitive to random fluctuations with larger subsets. 
This indicates that SGScore is both scalable and robust, providing consistent evaluations of factual consistency regardless of the dataset size.
\begin{table*}[t]
\centering
\definecolor{colorbg@intro}{RGB}{240,240,240} 
\begin{tikzpicture}[
    node distance=1cm and 0.5cm,
    auto,
    description/.style={
        rectangle,
        rounded corners=2pt,
        draw,
        thick,
        fill=colorbg@intro,
        text width=\textwidth,
        align=left,
        scale=1.0
    }
]
\node[description] {%
\textcolor{blue}{messages} = [\{ \texttt{"role"}: \texttt{"user"}, \texttt{"content"}: ``Given a file listing object categories, generate 3 realistic instances for each category by adding common-sense visual attributes such as color, size, material, or texture. Ensure that the attributes are contextually appropriate. For example, transform `apple' into `red apple', `green apple', and `ripe apple'; or `car' into `white car', `small car', and `luxury car'. The generated instances should be diverse and plausible.  formulate the output as a dict in JSON like \{``apple'': [``red apple'', ``green apple'', ``ripe apple''], ``car'':[``white car'', ``black car'', ``luxury car'']\}
\} ];
};
\end{tikzpicture}
\caption{Prompt for refining object categories with attributes. The  LLM receives  a list of object categories as input and outputs corresponding fine-grained object descriptions.
}
\label{tab:prompt_augment_objs}
\end{table*}

\subsection{Do We Really Need Attributes for SG2IM?}
\label{sec:supp_attr}
One potential concern for \emph{MegaSG} is its lack of attribute annotations, 
as users might expect attributes to play a role in scene generation—for example, distinguishing between \texttt{a red apple is above the black car} and \texttt{a black apple is above the red car}. 
Does omitting attributes in \emph{MegaSG} adversely affect the modeling of such distinctions?

Recent work~\cite{clark2023text} demonstrates that diffusion models can effectively bind attributes (\eg, color, texture, shape) under certain conditions. 
However, most existing SG2IM approaches~\cite{johnson2018image,yang2022diffusion,farshad2023scenegenie,shen2024sg,liu2024r3cd} do not explicitly model attributes, despite their availability in the Visual Genome dataset. 
To evaluate the impact of attribute modeling, we augment $775$ object categories in \emph{MegaSG} with attribute labels (\eg, \{\emph{apple} $\to$ \emph{red apple}, \emph{green apple}, \emph{ripe apple}\},
\{\emph{door} $\to$ \emph{wooden door}, \emph{red door},
\emph{glass door}\}, \etc 
). 
Each object category is augmented three times by a multimodal LLM
(the prompt can be found in \cref{tab:prompt_augment_objs}), 
yielding a total of $2,325$ text prompts for diffusion models. 
As shown in Table~\ref{tab:attr_objs}, diffusion models consistently handle both single-object and attribute-bound object generation with minimal performance differences, suggesting that explicit attribute modeling offers only marginal benefits in this setting.


More importantly, our experiments (\eg, \cref{tab:feedback_exp} and \cref{fig:complexity_impact}) reveal that current diffusion models struggle to capture spatial and interactive relationships, 
rather than merely modeling individual objects. 
This shortcoming can be attributed to two factors. 
First, T2I models typically employ a CLIP text encoder to process the textual input;
however, due to dataset biases (\eg, LAION 5b~\cite{schuhmann2022laion}) during training, the encoder learns a visual–text alignment that favors object presence over the relationships between objects. 
Similarly, the training data for Stable Diffusion models predominantly consists of single-object images, 
providing limited exposure to the alignment of multiple objects and their interactions.

Considering the trade-off between annotation cost and the minimal gains from introducing attributes, 
we opted to omit attributes in the construction of \emph{MegaSG}. 
Our focus is on effectively evaluating scene graph-to-image generation (SG2IM), 
thus providing a promising direction for optimizing diffusion models.

\begin{table}[t]
    \centering
    \begin{tabular}{l|cc}
        \toprule
         & SD v1.5 & SDXL \\
        \midrule
         \emph{w.o.} attributes& $93.90\pm0.29$  & $92.40\pm0.36$ \\
         \emph{w.} attributes &  $90.12\pm0.26$ & $93.34\pm0.15$ \\ 
        \bottomrule
    \end{tabular}
    \caption{Recall comparison of single-object vs.\ attribute-bound generation on 775 object categories.
    The recall evaluation is performed by Gemini 1.5 Flash.
    }
    \label{tab:attr_objs}
\end{table}

\subsection{Compared to Existing Benchmarks}
\label{sec:supp_benchmarks}
Although our work focuses on Scene Graph-to-Image (SG2IM) generation, textual inputs can be directly parsed into structured scene graphs. 
To validate the effectiveness of our proposed \emph{SGScore} in enhancing factual consistency, 
we compare \emph{SGScore} with two text-to-image (T2I) benchmarks, namely TIFA~\cite{hu2023tifa} and DSG~\cite{cho2024dsg}.

\begin{table*}[t]
\centering
\definecolor{colorbg@intro}{RGB}{240,240,240} 
\begin{tikzpicture}[
    node distance=1cm and 0.5cm,
    auto,
    description/.style={
        rectangle,
        rounded corners=2pt,
        draw,
        thick,
        fill=colorbg@intro,
        text width=0.985\textwidth,
        align=left,
        scale=1.0
    }
]
\node[description] {%
\textcolor{blue}{messages} = [\{ \texttt{"role"}: \texttt{"user"}, \texttt{"content"}: 
``Extract a structured scene graph from the following sentence. In this scene graph, include all physical objects even if they do not participate in any relationships (i.e., allow orphan nodes). The output should be a dictionary with two keys:
- ``objects'': a list of all physical objects mentioned in the sentence.
- ``relationships'': a list of dictionaries, each containing:
    - ``source'': the subject (a physical object) of the relationship.
    - ``target'': the object (a physical object) of the relationship.
    - ``relation'': the predicate describing the relationship between the source and target.
Ensure that both ``source'' and ``target'' are physical objects (not abstract concepts).

Sentence: ```\{input\_sentence\}'''

Output format:
\{
    ``objects'': [ ... ],
    ``relationships'': [
         {``source'': ``...'', ``target'': ``...'', ``relation'': ``...''},
         ...
    ]
\}

];};
\end{tikzpicture}
\caption{Prompt for extracting a textual scene graph from a caption.}
\label{tab:prompt_text_sg}
\end{table*}
\definecolor{cbRed}{RGB}{213,94,0}    
\definecolor{cbGreen}{RGB}{0,158,115}   
\begin{figure}[t]
    \centering
    \begin{tikzpicture}
      \begin{axis}[
        width=0.9\columnwidth,
        height=0.6\columnwidth,
        ylabel={Score},
        xmin=0.8, xmax=4.2,
        ymin=0.6, ymax=0.9,
        xtick={1,2,3,4},
        xticklabels={SD v1.5, SD v2.1, SDXL, SD3},
        grid=major,
        ymajorgrids=true,
        xmajorgrids=false,
        legend pos=south east,
        legend style={font=\small, draw=none, fill=none},
        tick label style={font=\small},
        label style={font=\small}
      ]
      
      \addplot[
        mark=square*,
        thick,
        color=cbRed,
        mark size=2pt
      ] coordinates {
        (1,0.7965)
        (2,0.8233)
        (3,0.8521)
        (4,0.8870)
      };
      \addlegendentry{TIFA}

      \addplot[
        mark=triangle*,
        thick,
        color=cbGreen,
        mark size=2pt
      ] coordinates {
        (1,0.6786)
        (2,0.7359)
        (3,0.7714)
        (4,0.8195)
      };
      \addlegendentry{SGScore}

      \end{axis}
    \end{tikzpicture}
    \caption{Comparison of TIFA and SGScore across different Stable Diffusion models on TIFA v1.0 benchmark.}
    \label{fig:comp_tifa_sgscore}
\end{figure}
TIFA~\cite{hu2023tifa} examines faithfulness in Visual Question Answering (VQA) by leveraging GPT-3~\cite{brown2020language} for question generation and using question answering modules such as mPLUG~\cite{li2022mplug} and BLIP-2~\cite{li2023blip2}. 
It also introduces the TIFA v1.0 benchmark, which comprises $4,081$ text prompts.
With these text prompts, we synthesize images using different diffusion models to compare \emph{SGScore} with the TIFA score, 
evaluated using the \texttt{mPLUG-large} model.
To compute \emph{SGScore}, we utilize an LLM (\eg, Gemini 1.5 Flash) with the prompt detailed in \cref{tab:prompt_text_sg} to extract a textual scene graph from a caption.
From \cref{fig:comp_tifa_sgscore}, we observe a consistent trend between the TIFA score and \emph{SGScore} across different diffusion models on the TIFA v1.0 benchmark, indicating that both metrics reliably reflect factual consistency. 
However, while TIFA is designed for text-to-image (T2I) generation, 
our \emph{SGScore} focuses on scene graph-to-image (SG2IM) generation.

Similary, DSG~\cite{cho2024dsg} decomposes the text prompt into a series of tuples (\eg, entities, attributes, relations) and generate corresponding questions for verifying the faithfulness of text-to-image generation. 
However, due to the high computational cost (see \cref{sec:comp_cost}), we opted not to report DSG scores against TIFA and \emph{SGScore}.

\subsection{Computation Cost}
\label{sec:comp_cost}
\textbf{Computation cost for evaluation pipeline.}
We benchmark 5,000 samples using Gemini 1.5 Flash to assess computation cost. 
On average, input and output token consumption are ${\sim} 4.7$M and ${\sim} 3.9$M, respectively, with a cost of ${\sim} 1.5$~USD for 5k samples (as of March 2025)\footnote{Gemini API pricing~\cite{google2024gemini}}. 
In contrast, DSG~\cite{cho2024dsg} requires ${\sim} 45$M input tokens and ${\sim} 0.7$M output tokens, resulting in a cost of ${\sim} 7$~USD for benchmarking 5k samples—even when using the considerably cheaper and more powerful LLM model GPT-4o mini~\cite{gpt4opricing} compared to the originally employed \texttt{gpt-3.5-turbo-16k-0613} in the paper.

\noindent\textbf{Computation cost for scene graph feedback.}
While scene graph feedback can be iteratively applied to refine generated results, 
we perform a single refinement step once discrepancies are detected by our evaluation pipeline. 
As shown in \cref{tab:benchmark_comparison_mega} and \cref{tab:benchmark_comparison_vg}, 
this approach yields a significant performance gain despite the additional computational overhead. 
A related work in T2I generation~\cite{wu2024self} also employs an iterative correction pipeline, integrating an object detector with an LLM, which requires extra computational cost.

\end{document}